\documentclass[letterpaper]{article}
\usepackage{iccc}
\usepackage{todonotes}
\usepackage{url}

\makeatletter
\g@addto@macro{\UrlBreaks}{\UrlOrds}
\makeatother

\usepackage{times}
\usepackage{helvet}
\usepackage{courier}

\usepackage{booktabs}
\usepackage[normalem]{ulem}
\useunder{\uline}{\ul}{}
\usepackage{longtable}
\usepackage{lineno}

\title{Towards Responsible AI Music:\\an Investigation of Trustworthy Features for Creative Systems}
\author{
Jacopo de Berardinis\textsuperscript{1}\thanks{Corresponding author: \texttt{jacodb@liverpool.ac.uk}},
Lorenzo Porcaro\textsuperscript{2}\thanks{Work done when the author was at European Commission's Joint Research Centre. The views expressed in this document are purely those of the authors and may not in any circumstances be regarded as stating an official position of the European Commission.},
Albert Meroño-Peñuela\textsuperscript{3},
Angelo Cangelosi\textsuperscript{4},
Tess Buckley\textsuperscript{5}\\
\textsuperscript{1} Department of Computer Science, University of Liverpool, UK\\
\textsuperscript{2} Department of Computer, Control and Management Engineering, Sapienza University of Rome, Italy\\
\textsuperscript{3} Department of Informatics, King's College London, UK\\
\textsuperscript{4} Department of Computer Science, University of Manchester, UK\\
\textsuperscript{5} Association of AI Ethicists
}

\newcommand{\RAIMw}{{\url{https://amresearchlab.github.io/raim-framework/}}}

\begin{document}
\maketitle
\begin{abstract}
\begin{quote}
Generative AI is radically changing the creative arts, by fundamentally transforming the way we create and interact with cultural artefacts.
While offering unprecedented opportunities for artistic expression and commercialisation, this technology also raises ethical, societal, and legal concerns.
Key among these are the potential displacement of human creativity, copyright infringement stemming from vast training datasets, and the lack of transparency, explainability, and fairness mechanisms.
As generative systems become pervasive in this domain, responsible design is crucial.
Whilst previous work has tackled isolated aspects of generative systems (e.g., transparency, evaluation, data), we take a comprehensive approach, grounding these efforts within the Ethics Guidelines for Trustworthy Artificial Intelligence produced by the High-Level Expert Group on AI appointed by the European Commission -- a framework for designing responsible AI systems across seven macro requirements.
Focusing on generative music AI, we illustrate how these requirements can be contextualised for the field, addressing trustworthiness across multiple dimensions and integrating insights from the existing literature.
We further propose a roadmap for operationalising these contextualised requirements, emphasising interdisciplinary collaboration and stakeholder engagement. 
Our work provides a foundation for designing and evaluating responsible music generation systems, calling for collaboration among AI experts, ethicists, legal scholars, and artists.
This manuscript is accompanied by a website: \RAIMw.
\end{quote}
\end{abstract}

\section{Introduction}\label{sec:introduction}

Generative AI is fundamentally transforming the artistic landscape, demonstrating remarkable capabilities across various modalities and domains.
From visually stunning images generated by diffusion models \cite{rombach2022high} to the literary fluency of large language models \cite{touvron2023llama}, generative models continue to expand the boundaries of what we consider creative expression \cite{henrik2023genai}.

The music domain has been also impacted, with generative models achieving impressive results on both symbolic and audio music \cite{briot2020deep,copet2024simple}.
The current variety of music models is broad and diversified, and has already enabled new forms of artistic co-creation \cite{huang2020ai}.
These range from the automatic generation, completion, and alteration of chord progressions and melodies, to the creation of mashups and audio snippets from textual prompts \cite{agostinelli2023musiclm}.
Due to their success, some of these systems have already been integrated into commercial software, allowing users to generate music pieces from their desiderata.

The rise of AI music also raises profound ethical concerns, especially when access to generative systems and fruition of their content are made commercially available.
Generative AI relies on the creative work of artists for its success -- yet currently offers no compensation for their implicit data contributions \cite{samuelson2023generative}.
Paradoxically, the technology's potential to automate music creation threatens to displace the very artists whose work made it possible.
While this goes beyond the augmenting artistic possibilities \cite{sturm2019artificial}, it can also pave the way to highly lucrative business opportunities, given the low cost of non-human musicians and their full control \cite{morreale2021mai}.

Generative music AI also brings profound social and cultural implications.
While promising to democratise music creation by lowering barriers to entry, it also threatens to disrupt the livelihoods of human musicians, necessitating reskilling and upskilling initiatives to mitigate potential displacement \cite{henrik2023genai}.
This disruption raises concerns about the long-term impact on human creativity, with the potential for cognitive atrophy as AI systems increasingly assume the role of composer and performer \cite{saetra2019ghost}.
Culturally, the emergence of new genres and styles fuelled by AI could enrich the musical landscape, pushing creative boundaries.
However, this evolution must be carefully steered to avoid the pitfalls of bias.
Current models often exhibit a Western-centric focus in their training data, risking the marginalisation of non-Western musical traditions and hindering cultural diversity \cite{barnett2023ethical}.
Furthermore, the ability to generate music mimicking deceased artists raises ethical concerns about artistic integrity, authenticity, and the potential for misrepresentation or exploitation.

From a technical perspective, generative models that fully learn to compose from data by maximising a general, domain-agnostic objective (e.g., predicting the next token) are also criticised for the lack of transparency and explainability.
At the data level, this is related to the challenge of keeping track of where the model derives its musical ideas. 
The lack of transparent source attribution in the music metadata may prevent giving recognition to those artists who contributed their music as training material, leading to immediate implications for copyright and revenue sharing \cite{drott2021copyright,sturm2019artificial}.
Similarly, the lack of explainability represents a technological barrier for end-users, as there is little or no understanding of the creative process underneath.
Explainability is a desirable component for generative systems, as it facilitates the interaction with artists, and particularly, the ability to control/steer the system based on domain knowledge \cite{bryan2022exploring}.
Finally, the ``creative space'' learned by data-driven systems may also lack of musical plausibility \cite{guo2022domain}, meaning that solutions generated from such models may violate common notions of music theory.
While these deviations can be creatively stimulating, they often create a barrier to effective communication and collaboration between music theorists and AI developers.

In sum, the majority of generative music systems cannot yet be deemed responsible by design, which raises serious concerns related to their large-scale adoption.
Ultimately, navigating the rise of generative music AI requires a delicate balance: fostering innovation while safeguarding the artistic integrity and economic sustainability of human creators, ensuring that AI serves to augment, not replace, the irreplaceable human element at the heart of music.

\subsection{Responsible AI initiatives and projects}

In response to these challenges, numerous projects and initiatives are advocating for the responsible development and use of Music AI.
These efforts focus on various aspects, requirements, and dimensions of responsible AI, aiming to ensure that AI technologies in music are developed and employed ethically and beneficially.
We discuss hereafter few examples among the most recent efforts, and we point to the Music AI Ethics tracker\footnote{\url{https://www.waterandmusic.com/data/ai-ethics-tracker}} maintained by Water\&Music for an updated overview of related initiatives.

The ``\textit{Human Artistry Campaign}'' was established in 2023 as a collaborative effort to advocate for responsible AI development within the creative sector\footnote{\url{https://www.humanartistrycampaign.com}}.
Started as a coalition of 150+ organisations, including major music industry bodies (IFPI, RIAA, BPI) and artist representative groups (AIM, Featured Artists Coalition, Impala), the Campaign has articulated seven core principles to guide AI's integration with creative practices.
These principles underscore the continued primacy of human expression and the necessity of respecting copyright and intellectual property rights in AI development.
Furthermore, they emphasise transparency in AI training processes and the inclusion of creators in policy discussions surrounding AI.
These principles assert that: 1) AI should function as a tool to support, not replace, human creativity; 2) human-created works remain essential to cultural expression; 3) the use of copyrighted works in AI development requires proper authorisation and licensing; 4) governments should not grant AI developers exemptions from copyright law; 5) copyright protection should apply exclusively to human-generated works; 6) transparency in AI development and the use of training data is crucial; and 7) creators must be actively involved in shaping AI policy.
The Campaign's formation signals a proactive effort by the creative community seeking to ensure that AI technologies serve to augment, rather than undermine, human artistry.

Recently, Roland Corporation and Universal Music Group spearheaded an initiative called ``\textit{AI for Music}'', introducing the ``\textit{Principles for Music Creation with AI}''\footnote{\url{https://aiformusic.info}}.
This campaign, endorsed by over 50 organisations including NAMM, BandLab Technologies, and Splice, seeks to guide the ethical development and implementation of AI in music.
The principles emphasise: 1) the centrality of music to human wellbeing, 2) the inseparability of human creativity and meaningful music, 3) AI's potential to amplify human expression, 4) the necessity of protecting human-created works and artists' rights, 5) transparency as a driver to responsible and trustworthy use and development of AI, 6) respecting the perspectives of human creators, and 7) the commitment to supporting human creativity through technology.
This initiative reflects a growing consensus within the music industry on the importance of responsible AI integration that prioritises human expression and artistic integrity.

UK Music, a British umbrella organisation representing the collective interests of the UK's commercial music industry, released a policy position paper on AI \cite{ukmusic2023white} and a manifesto \cite{ukmusic2023manifesto} outlining its stance on the use of AI in music.
This paper emphasises the importance of ensuring AI serves to support human creativity in music.
It calls for policymakers to implement the following measures: 1) mandate the identification of AI-generated music through labelling or metadata, 2) uphold creators' and rights-holders' control over the use of their works, 3) require record-keeping of ingested works in AI training processes, and 4) establish personality/image rights within the UK legal framework.
Representing stakeholders including artists, record labels, and music publishers, and with key members such as the Musicians' Union (MU), the British Phonographic Industry (BPI), and PRS for Music, UK Music actively collaborates with the UK government on policy matters.
Overall, these recommendations aim to protect human artistry, ensure proper attribution, and prevent the misuse of AI in music creation.

\textbf{University-led projects} have also emerged, often focusing on specific issues such as bias in training data and the impact of AI on the music industry.
For example, the \textit{Music RAI} project \footnote{\url{https://music-rai.github.io}} tackles the issue of bias in AI music generation, and is driven by the Western bias of current AI music models, which can lead to the marginalisation of non-Western musical traditions.
By building an international community of researchers, musicians, and AI experts, this project aims to create and share resources that promote the inclusion of marginalised musical genres in AI music creation.
This involves establishing diverse musical datasets and developing accessible AI tools for artists.

Similarly, the \textit{AI:OK}\footnote{\url{https://ai-ok.org}} project,
lead by Dublin City University and funded by the Irish government, seeks to establish a trustmark to identify music created ethically using AI to protect the rights of music stakeholders.
AI:OK focuses on self-regulation, IP protection, transparent standards, and promoting innovation in the music industry.
The project is supported by key industry players and research institutions, including Enterprise Ireland, Universal Music Group, and the Insight SFI Research Centre for Data Analytics.

\subsection{Our contribution}

Motivated by the aforementioned concerns surrounding music generative systems, and building upon parallel works in the literature, here we attempt to unify and complement these efforts within the general framework of Trustworthy AI\footnote{Note that \textit{trustworthiness} subsumes \textit{trust} in AI, as it encompasses a broader set of requirements beyond just trust, including aspects like reliability, safety, and accountability \cite{Vereschak2021}.} produced by \citeauthor{ala2020assessment}.
Our work is based on the assumption that generative systems can be designed and evaluated for trustworthiness, and as such, they should account for responsible features and mechanisms that are central for human use and collaboration.
We posit that ``thinking in trustworthiness'' in generative systems is a proxy to fostering a creative and rightful relationship between humans and AI -- in accordance with ethical, societal, legal, and scientific principles.

Following a review of relevant background concepts, this article is structured in two main parts.
First, we contextualise each of the 7 requirements for Trustworthy AI to the domain of Generative Music AI, drawing parallels with the literature, highlighting the main challenges, and contributing to the definition of 45 responsible features.
Second, we outline a roadmap for operationalising these features, emphasising the crucial role of interdisciplinary collaboration, stakeholder engagement, and an anticipatory approach to address the rapidly evolving ethical landscape of AI music.

\section{Background}\label{sec:framework}

Trustworthy AI encompasses systems designed and implemented to adhere to fundamental ethical principles, technical robustness, and legal compliance.
Given their sensitive design and their compliance to specific requirements, trustworthy systems are often implemented in critical applications where privacy, safety, societal well-being, or ethics are of primary concern \cite{floridi2019establishing}.

Trustworthy AI is an emergent field driven by academic research, industrial cooperation, and support from institutional bodies.
In the EU, synergies between these parties have produced referential outputs, for example, the \textit{Ethics Guidelines for Trustworthy Artificial Intelligence} \shortcite{hleg2019ethics}(AI HLEG), a document prepared by the High-Level Expert Group on Artificial Intelligence -- an independent expert group appointed by the European Commission in 2018.
The guidelines include \textbf{7 macro-requirements} that AI systems should meet to be trustworthy: 1) human agency and oversight; 2) robustness and safety; 3) privacy and data governance; 4) transparency; 5) diversity, non-discrimination, fairness; 6) societal and environmental well-being; 7) accountability.
These requirements are of general applicability and relate to different stakeholders in the systems' life cycle (developers, deployers, end-users, broader society).
Following a piloting process, the guidelines lead to the creation of the \textit{Assessment list for Trustworthy AI (ALTAI)} \shortcite{ala2020assessment}.

\subsection{Related Trustworthy AI frameworks}

The EU's AI Act \cite{EU_AI_Act} came into force in 2024, providing a legal framework for regulating AI systems.
This legislation adopts a risk-based pyramid approach, categorising AI systems into four risk levels: \textit{unacceptable}, \textit{high}, \textit{limited}, and \textit{minimal}.
Notably, the AI Act also addresses General Purpose AI (GPAI) models, mandating technical documentation, copyright compliance, and public summaries of training data.
GPAI models identified as posing systemic risks are subject to further obligations, including model evaluations, incident reporting, and cybersecurity measures.
The AI Act's extraterritorial scope impacts global providers whose systems are used within the EU, setting a precedent for responsible AI development worldwide.

Recent endeavours have highlighted several challenges into the applicability of legal framework to Generative AI.
\citeauthor{helberger2023chatgpt} argue that the EU AI Act's risk-based approach is challenged by the unique nature of generative AI, specifically its dynamic context and vast scale of use.
They highlight the difficulty of categorising such systems into the risk categories, the unpredictability of future risks stemming from their widespread use, and the potential for private risk ordering to undermine the Act's goals.
Instead, they propose considering generative AI as a separate risk category, subject to dynamic risk monitoring and regulatory scrutiny of user-provider contracts.

Similarly, \citeauthor{novelli2024aiagen}, discuss the AI Act's shortcomings in addressing the unique challenges posed by Generative AI.
The Act's risk-based categorisation may struggle to capture the emergent behaviour of LLMs and their potential for misuse.
GDPR compliance for Generative AI is complicated by vast training datasets and the potential for data leakage, requiring careful examination of legal bases for data processing and the right to erasure.
Copyright issues arise from the use of copyrighted material in training data and the unclear IP status of AI-generated outputs.
Finally, cybersecurity vulnerabilities, including adversarial attacks and misinformation, necessitate robust safeguards and potential revisions to existing legislation like the Cyber Resilience Act (CRA) and the Digital Services Act (DSA).

Overall, the EU AI Act, while groundbreaking, represents a legal framework, whereas the AI HLEG's Ethics Guidelines constitute an ethical framework.
While the former draws inspiration from the latter, the AI Act does not supersede these ethical guidelines: the Act focuses primarily on risk mitigation and legal compliance, while the Guidelines delve deeper into the ethical dimensions of AI development and deployment.
Therefore, our work takes a step back and begins with the Ethics Guidelines for Trustworthy AI, using them as a foundation for analysing and addressing the ethical challenges posed by Generative Music AI.
This approach ensures that our analysis also considers the broader societal and ethical implications of this technology.

Beyond Europe, the landscape of Trustworthy AI is populated by several influential works.
Key among these is the international standard ISO/IEC 42001 \cite{ISO_IEC_42001_2023}, which provides requirements for establishing, implementing, maintaining, and continually improving an AI management system.
Other notable frameworks include the OECD \textit{Principles on AI}, which emphasise human-centric values and have garnered broad international support \cite{oecd2019oecd}, and the G20 \textit{AI Principles}, which further underscore the importance of inclusivity and sustainable development in the context of AI \cite{g202019g20}.
The UNESCO \textit{Recommendation on the Ethics of Artificial Intelligence} stands out for its global consensus, addressing a wide range of ethical, social, and cultural impacts \cite{unesco2021recommendation}.
Additionally, industry-led initiatives like the \textit{Partnership on AI} and the \textit{IEEE Global Initiative on Ethics of Autonomous and Intelligent Systems} provide crucial guidance on best practices and ethical considerations from a multi-stakeholder perspective \cite{partnership2016partnership,ieee2019ethically}.
At a national level, frameworks such as the US \textit{NIST Risk Management Framework} \cite{NIST_RMF_2025} and Singapore's \textit{Model AI Governance Framework} offer valuable insights into practical implementation and industry-specific considerations \cite{singapore2020model}.
These contributions are often accompanied by practical supporting tools to facilitate the implementation of these frameworks.
For example, the UK's AI Management Essentials tool \cite{DSIT_AIME_2024} is a self-assessment tool for organisations to implement responsible AI management, drawing upon leading frameworks such as the EU AI Act, NIST, and ISO/IEC 42001; and the EU AI Code of Practice \cite{EU_AI_Office_GPAI_Code_2025} will guide the operationalisation of the AI Act.

While these other frameworks contribute significantly to
the broader discourse on AI ethics, we found the HLEG guidelines to provide a flexible, granular and actionable framework that is better suited for translating ethical principles into concrete design features for Responsible AI music.

Nevertheless, while these various frameworks aim for broad cooperation, the current tense geopolitical landscape presents significant challenges to achieving unified global AI governance.
There is considerable uncertainty surrounding the future direction of AI policy, as illustrated by shifts in US national priorities, including the removal of initiatives such as the \textit{Blueprint for an AI Bill of Rights} \cite{whitehouse2022blueprint} in 2025.
The themes of AI nationalism and acceleration are increasingly evident, as demonstrated by both the UK and US declining to sign the Declaration following the AI Action Summit in France \cite{KleinmanMcMahon2025AI}.
This suggests a potential shift away from coordinated international governance approaches towards more nationalistic AI development policies.
Despite these challenges, the ongoing development of AI ethics frameworks demonstrate a continued commitment to responsible innovation.

\begin{figure*}[t]
    \centering
    \includegraphics[width=.95\linewidth]{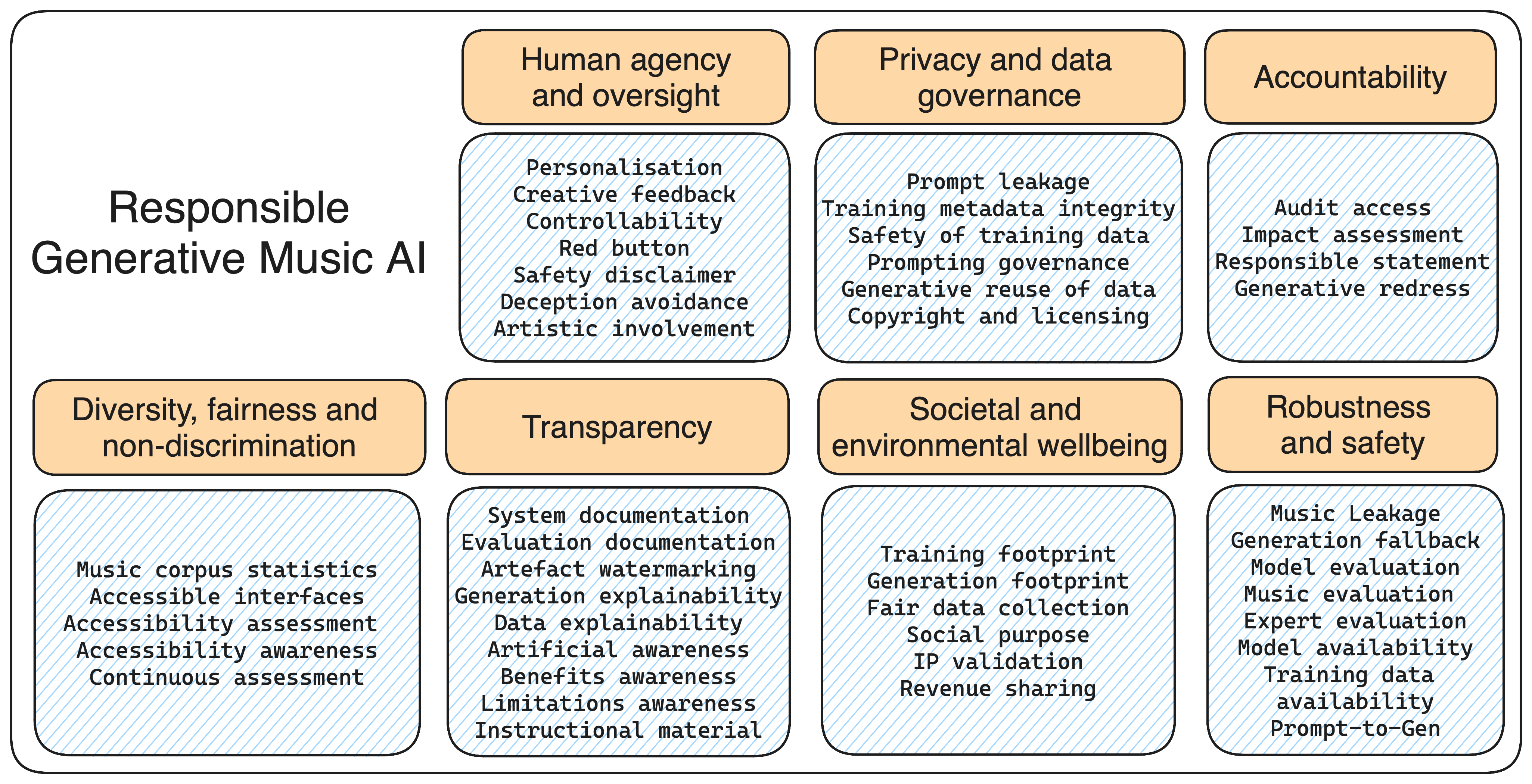}
    \caption{Overview of the Trustworthy AI macro-requirements (\textit{orange}) along with their contextualisation in Generative Music AI (\textit{blue}). Each features is described in Section~\ref{sec:framework}2 with and a summary is provided in Table~\ref{tab:features-all-c} (Appendix).}
    \label{fig:features-diag}
\end{figure*}

\subsection{Music models and systems}
Our contextualisation of the Trustworthy AI framework to the domain of Generative AI Music, aims to identify responsible features that can drive the design and the evaluation of generative tools.
For this pursuit, a distinction needs to be made between \textit{music model} and \textit{generative system}.
\begin{itemize}
    \item A music \textbf{model} can be defined as an algorithmic procedure that either encodes a set of rules explicitly, or learns them from the data and the task it is provided.
    These rules, e.g., a probability distribution for predicting the next note or chord in a piece, or a set of logical statements, can be used to generate, complete, or manipulate music.
    \item  A generative music \textbf{system} encompasses the whole computational infrastructure built on top of a music model to enable users to interact with the model and make use of its outputs, without the need of its inner workings.
    This includes both technical and the regulatory aspects, such as the interface, the logic which abstract or hide certain parameters of the model, the way the model's predictions are consumed; but also the data management system, the legal framework regulating the exchange of data, etc.
\end{itemize}

Typically, a generative system is implemented in such a way as to conveniently wrap the functionalities of a particular model, meaning that a model can provide the computational backbone to various generative systems (e.g., plugins for music editors, production environments, smart instruments).
For example, MusicVAE \cite{roberts2018hierarchical} has been reused in different applications, such as Beat Blender, Melody Mixer, Latent Loops, and is also available through Magenta Studio -- a plugin for the DAW Ableton Live.

The distinction between model and system is a peculiar aspect to Generative AI, as their design and implementation typically involve different stakeholders, such as machine learning engineers or mathematicians for the former, and UX designers, software developers, data engineers for the latter, but also share music experts as a common denominator driving the evaluation efforts.

\section{Trustworthiness in Generative Music AI}

With this distinction in mind, we discuss trustworthiness in Generative Music AI by defining\textbf{ 45 responsible features} that expand and contextualise the aforementioned macro-requirements to this domain.
These are illustrated in Figure~\ref{fig:features-diag} and outlined in Table~\ref{tab:features-all-c} (Appendix), with the following subsections providing a systematic overview into their definition, motivation, and potential implications.
To facilitate the dissemination, evaluation, and evolution of these features, we have launched a dedicated website at \RAIMw.

The responsible features presented in this work were identified through a multi-faceted approach.
First, we conducted a review of existing literature on Trustworthy AI (in relation to the EU's AI HLEG's Ethics Guidelines), and Generative AI, with a particular focus on music generation and music technology.
This allowed us to identify specific aspects, concepts, and methodologies already addressed in these fields that could be related to the dimensions and requirements of trustworthy AI.
Second, we drew upon insights, critical points, and perspectives that emerged from interview sessions with music artists and ethical experts.
These interviews provided valuable qualitative data on the ethical implications of Generative Music AI from the viewpoints of those directly involved in its creation and use.
Finally, we considered the principles and best practices set forth by the Responsible AI initiatives outlined in the introduction.
Overall, this approach aims to ensure that the proposed feature set, while not exhaustive, serves as a preliminary foundation for initiating a broader discussion among diverse music stakeholders (including Music AI experts, music artists, ethicists, legal experts, and music listeners) to collectively rank the importance of these features and inform the design of trustworthy-by-design AI music models and systems.

These responsible features can be categorised according to the seven key requirements of trustworthy AI as follows.

\begin{enumerate}
    \item \textbf{Human Agency and Oversight}. This requirement ensures that users maintain control over AI systems and their outputs. This translates to features that support user understanding of the system's capabilities and limitations, offer control over the generative process (e.g., through interactive feedback and multi-modal conditioning), and prioritise user safety by providing safeguards against potentially harmful or inappropriate musical outputs.
    \item \textbf{Robustness and Safety}.  This focuses on the reliability and security of generative music AI models and systems. Key features include a systematic evaluation of the model's accuracy using diverse methods (music modelling, statistical comparisons, composition evaluation, and listening tests), ensuring reproducibility through the availability of models, code, and training data, and protecting against attacks that could leak copyrighted material or compromise the system's integrity once deployed.
    \item \textbf{Privacy and Data Governance}. The focus of this requirement is on responsible data handling throughout the lifecycle of a generative music AI. This includes protecting user privacy by preventing data misuse and discrimination, implementing data access protocols, and ensuring the quality and integrity of training data, particularly music metadata. It also necessitates addressing copyright and licensing issues by documenting data sources, complying with relevant regulations, and providing clear information on the copyright of generated outputs.
    \item \textbf{Transparency}. This requirement promotes explainability and openness in generative music AI. Traceability is achieved through comprehensive documentation of the model's design, evaluation, and datasets, along with robust watermarking of AI-generated music (system-wise). Explainability features enable users to understand the generation process and trace outputs back to their source material. Clear communication about the system's purpose, capabilities, limitations, and potential musical biases is essential for informed use.
    \item \textbf{Diversity, Non-discrimination and Fairness}. This requirement focuses on promoting inclusivity and avoiding bias in generative music AI. Key features include mitigating unfair bias in training data and model outputs, designing accessible interfaces for diverse users, and actively involving stakeholders, especially musicians, in the design, development, and evaluation processes.
    \item \textbf{Societal and Environmental Well-being}. This requirement highlights the broader impact of generative music AI on society and the environment. It encourages the development of sustainable and environmentally friendly AI by promoting awareness of the environmental footprint. It also necessitates considering the socioeconomic impact of these systems, including potential effects on employment, cultural diversity, and the revenue processes.
    \item \textbf{Accountability}. This requirement ensures responsible development and deployment of generative music AI systems. It includes provisions for auditability, allowing internal and external assessments of the system's compliance with responsible AI principles. It also necessitates identifying and mitigating potential negative impacts, making ethical trade-offs when necessary, and redress mechanisms for any unjust adverse impacts.
\end{enumerate}

The following subsections provide a systematic overview of the features, their definition, motivation, and implications.

\subsection{1) Human Agency and Oversight}

Central to the design of trustworthy systems is the capacity of individuals to understand and meaningfully influence the actions of AI systems (\textit{human agency}) while monitoring and interacting their behaviour to operate responsibly and align with human values and preferences (\textit{human oversight}).

\subsubsection{Human agency}
refers to the ability of users to retain power over their choices, understand and question the system, and ensure it works in support of their goals rather than manipulating them.
Overall, it mandates that the system's design is centred around the user, who should be made aware of any deceptive behaviour or potentially harmful outcomes.

The requirement is highly relevant to Generative AI, which is known to have persuasive power \cite{wach2023dark}.
Critical judgement and control are necessary especially when the outputs could produce legal effects, which in this case, may relate to the copyright and ownership of the generated material.
A conversational generative system, for example, should not attempt to claim ownership or novelty of the generations when the music is possibly plagiarised.
In relation to potential harmful impacts, we also expect a system to provide appropriate safety disclaimers if it can generate music that can affect listeners or may be deemed inappropriate by some categories of users.
For example, lyrics may contain offensive or toxic language, given the unsafe behaviour of current language models \cite{baheti2021just} and their increasing use in the music domain \cite{ma2021ai}.
The same applies to systems that continually generate music with extreme or potentially disturbing acoustic characteristics, which has demonstrable effects on listeners' emotional state \cite{di2022consonance}, and may elicit unpleasant judgements \cite{koelsch2006investigating}.

Another feature relating to human agency is \textit{personalisation} -- the ability of a model to use the musical preferences or the repertoire provided by the artist to personalise the style of the generations.
Common ways to achieve this include enforcing musical properties \cite{mira2023automated}, style transfer  \cite{cífka2021style}, and tuning pre-trained music models to the provided repertoire \cite{mosbach2023few}.
The effectiveness of repertoire-based personalisation, however, is often limited by the dataset's size and diversity.
This may result in the model replicating existing works rather than generating more original music that adheres to the artist's signature style and sound.

\subsubsection{Human oversight}
encompasses all the mechanisms ensuring that users can actively control the behaviour of the system.
These include human-in-the-loop (HITL), human-on-the-loop (HOTL), and human-in-command (HIC).
HITL refers to the capability for human intervention in every decision cycle of the system, whereas HOTL relies on human feedback to improve its performance over time.
HIC refers to the capability to oversee the overall activity of the system.

In music generation, human-in-the-loop can be implemented in various ways.
Creative feedback can be used to guide and iteratively refine the generation process to align the outputs to the user's desiderata \cite{zhang2024musicmagus}.
Besides language \cite{copet2024simple}, controllability can also be achieved by conditioning the model's generation using one or more modalities, such as a priming melody \cite{huang2018music}, a supporting harmonic progression \cite{brunner2017jambot}, a rhythmic sequence \cite{lattner2019high}, or emotion profiles \cite{bao2023emogen}.
Instead, HOTL calls for the involvement of creative professionals during the design process, or to provide feedback and experiences of use as part of a collaborative effort \cite{huang2020ai}.
Finally, both agency and oversight advocate for ``red buttons'' to halt the generation process if the music contains content that promotes violence, hate speech, or discrimination.

It is crucial to distinguish between human artistic expression, which often uses provocative language for social commentary or stylistic effect, and AI-generated content, which lacks the same contextual understanding and intent.
While we acknowledge the importance of preserving artistic freedom and cultural expression across all genres, AI systems should be designed to avoid causing harm or promoting harmful ideologies.
Certain types of content, such as threats or hate speech, may also have legal implications depending on the jurisdiction.
Therefore, we advocate for a balanced approach that prioritises user control and contextual awareness.
This includes implementing user controls that allow individuals to set their own content preferences and thresholds, potentially based on age or other factors.
Furthermore, it necessitates working with diverse artists and cultural experts during the design and development process to ensure the system respects various forms of artistic expression while avoiding the generation of genuinely harmful content.
The goal is not to censor artistic expression, but to prevent AI from generating content that could be used to incite hatred, violence, or discrimination for certain users.

\subsection{2) Robustness and Safety}

This requirement focuses on accuracy, reliability and reproducibility, resilience to attack, security, and fallback plans.

\subsubsection{Accuracy}
relates to the model's ability to make correct predictions and recommendations.
This is measured through appropriate evaluation methods that vary depending on the task and the desired qualities of the model's outputs.

While accuracy in music generation may be of debatable utility for creatives, evaluation is a crucial step in the development of any generative music model.
Accuracy in generative music AI refers to the model's ability to generate musical outputs that meet the user's expectations and adhere to desired musical qualities.
This is measured through appropriate evaluation methods that vary depending on the task and the creative goals behind the music generation process.
Diverse evaluation methods exist, focusing on factors such as adherence to musical rules (e.g., pitch range, rhythmic consistency) and subjective assessments of the listening experience.
These methods can be broadly categorised as follows \cite{carnovalini2020computational}.

\begin{description}
\item{\textit{Music modelling evaluation}}. To generate music, a model is often trained on a pretext task, such as predicting the next musical token in a piece (e.g., a note, chord, or a fragment of them) given past context.
The main assumption is that a model accurately predicting music, can potentially encapsulate notions of music composition.
Evaluating the predictive performance of a music model is commonly done via machine learning metrics, e.g., \textit{log-likelihood}, \textit{frame-level accuracy} \cite{boulanger2012modeling}, F-measure, precision, recall and perplexity \cite{ycart2017study}.

\item{\textit{Statistical comparisons}}. These methods compute a set of descriptive statistics on a sample of generations, so that they can be compared with those extracted from the training data. Examples of these statistics are \textit{pitch and note counts}, \textit{pitch class and note length histograms}, and \textit{average pitch interval}. Statistical comparisons provide a weak measure of the similarity between generated and training material \cite{yang2020evaluation}, which is often a proxy to detecting novelty and plagiarism \cite{sturm2016music}.

\item{\textit{Composition evaluation}}. The purpose of this evaluation is to formally assess generated pieces for certain musical properties or for theoretical plausibility. This can be done via computational measures \cite{chuan2018modeling,de2022measuring}, or by involving music experts \cite{sturm2017taking}. Given the scarcity of computational measures that can automate this process, the latter is still the prominent approach. This requires high expertise, involves subjectivity at different levels, and may not be accessible to anyone.

\item{\textit{Listening tests}}. These encompass both \textit{Turing tests} and \textit{blind comparisons}. In a Turing test, listeners attempt to distinguish between human-composed and computer-generated music. Despite their notoriety, Turing tests have been criticised in recent decades for their inherent biases and challenges in designing reliable experiments \cite{ariza2009interrogator,pease2011impact,yang2020evaluation}.  Blind comparisons offer an alternative by having listeners rank music generated by different systems based on high-level qualities like likeability and originality.
\end{description}

In contrast to natural language generation \cite{gehrmann2022gemv2}, music generation still lacks standardised frameworks and benchmarks.
Nonetheless, a music model is expected to provide a comprehensive evaluation to shed light on the musical properties of its generations, their tendency to plagiarise the training data, and ideally involve creative professionals or music experts in this process.
To facilitate comparison across models and systems, evaluation methods should align with established practices in the field.

Evaluation in the context of music generation challenges often involves a multifaceted approach, incorporating both objective and subjective measures.
For example, the AI Music Generation Challenge 2022 \cite{Sturm2023ai}, focusing on generating Irish reels, employed a three-stage evaluation process.
Initially, tunes were screened for plagiarism, rhythmic accuracy, and adherence to modal conventions.
Subsequently, expert judges rated the remaining tunes based on structure and melody.
Finally, a discussion among judges determined prize-worthy tunes.
Similarly, the AI Song Contest\footnote{\url{https://www.aisongcontest.com}}, an international competition for AI-composed songs, employs a jury to assess the use of AI in the songwriting process, while the public evaluates the overall quality of the song.
This combination of expert and public feedback provides a comprehensive assessment of both the technical and aesthetic aspects of AI-generated music.
These diverse evaluation approaches highlight the ongoing effort to establish standardized frameworks and benchmarks for assessing the quality and impact of generative music AI systems.

\subsubsection{Reliability and reproducibility} ensure that the results of a music model can be reproduced by third parties.
These are not merely technical considerations in generative music AI; they are relevant aspects that may impact legal, artistic, and cultural dimensions.
While perfect reproducibility might not always be a primary goal in creative endeavours, a reasonable degree of reliability and the ability to reproduce results are crucial for several reasons.

First, for AI developers, reliability and reproducibility are essential for evaluating and comparing different music models and systems (potentially reproducing the accuracy measures mentioned before).
This research-oriented goal allows for meaningful benchmarking and progress in the field.

Second, from a legal perspective, reliability and reproducibility play a vital role in establishing ownership and addressing copyright concerns.
In copyright disputes, the ability to demonstrate the precise process of music creation, including the specific model, parameters, and data used, can be critical in proving originality and differentiating between coincidental similarities and actual infringements.
This is akin to a composer demonstrating their creative process on a traditional instrument.
Furthermore, clear delineation of AI-generated components within a musical work is necessary for proper licensing and royalty distribution.

Third, from an artistic perspective, a degree of reliability is crucial for artists who wish to use generative AI tools as part of their creative process.
Just as a musician relies on the consistent behaviour of their instrument and tools, an artist using AI needs to be able to achieve intended musical outcomes with a reasonable degree of predictability.
While the ideal level of predictability is contextual and subjective, this facilitates true human-AI collaboration and control.

Finally, from a cultural perspective, reproducible systems contribute to the documentation of innovation and the evolution of music technology.
The ability to understand and replicate the processes behind AI-generated music will be important for future musicologists and for understanding the cultural impact of AI on music.

To implement this requirement, AI developers should provide all the necessary material to reproduce and facilitate the evaluation of their model.
This includes releasing, either publicly or on request, instructions, code, and model checkpoints (\textit{model availability}), in addition to the full training material (\textit{training data availability}).
Moreover, if a sample of generations is used in the evaluation, the model should be seeded and prompted to recreate the same musical content.
This reproducibility can be facilitated by sharing prompts and generation settings, for example through collaborative prompting approaches \cite{merono2025kg}.

\subsubsection{Resilience to attack and security}
protects against vulnerabilities that can allow systems to be exploited by adversaries targeting the data (data poisoning), the model (model leakage), or the hardware/software infrastructure.

For music AI, the leakage of copyrighted training material can be seen as an instance of model leakage \cite{sun2021adversarial}.
However, the motivation behind such attacks often goes beyond simply obtaining the original audio files, which are typically readily available through other means (e.g., streaming services).
Instead, a key concern is that a compromised model might be manipulated to generate outputs that are substantially similar to, or even direct reproductions of, copyrighted material without triggering copyright detection mechanisms.
This could allow an attacker to claim the generated music as original, potentially benefiting from licensing schemes offered by the generative AI platform (e.g., claiming full copyright ownership under a subscription plan).
In essence, the attacker leverages the model's vulnerability to bypass copyright restrictions and potentially profit from the unauthorised use of this material.

Furthermore, the increasing trend of using generative model outputs as training data for other models \cite{villalobos2022will} exacerbates this risk.
If leaked, copyrighted material, disguised as novel AI-generated music, is incorporated into the training data of a subsequent model, the copyright infringement becomes further propagated and harder to trace.
This creates a cascading effect, potentially leading to widespread, unintentional copyright violations across multiple AI systems.
Therefore, we expect a resilient generative system to not allow for the full or partial reconstruction of the training music material unless this is explicitly acknowledged and allowed by the copyright holders of the datasets.

\subsubsection{Fallback plan and general safety} 
mechanisms provide safeguards in case of problems that manifest during the functioning of the system.
For generative music AI systems, this means having procedures in place to handle situations where the generated output is undesirable, unsafe, or otherwise problematic for the target audience (e.g. a system for affective music generation for mood regulation).
A potential fallback plan for a system that needs to continually generate music, such as a background music system for a public space or an interactive music co-creation tool, is switching to a simpler generative strategy (e.g., a rule-based mode) whenever its content is considered harmful or inappropriate by listeners or exceeds pre-defined safety thresholds.
These thresholds might relate to musical characteristics (e.g., excessive complexity or abrupt changes), lyrical content (e.g., detection of hate speech), or potential copyright infringement (e.g., high level of similarity to existing works).

Beyond switching to a simpler generation mode, other fallback options could include temporarily muting the output and alerting a human operator, switching to a pre-approved playlist, or, in interactive systems, prompting the user for further guidance.
The specific fallback mechanism and the thresholds that trigger it should be carefully designed and tailored to the specific application and context, balancing creative freedom with the need to ensure a safe and positive user experience as part of the generation.

\subsection{3) Privacy and data governance}

Central to the goal of preventing unintended harm and achieving trust with users, data governance must encompass the quality and integrity of the data utilises, its pertinence within the intended domain of AI deployment, strict access protocols, and mechanisms to safeguard privacy.

\subsubsection{Privacy and data protection}
mechanisms should be embedded in any process handling information from users.
This is crucial as AI can infer sensitive personal information like age, gender, sexual orientation, preferences and beliefs.

Such information could permeate from the users' interactions with a generative system (e.g., textual prompts, audio recordings, a video for soundtrack generation), or stem from the training material (e.g., a listening experience).
Strong safeguards should prevent data misuse and discrimination by ensuring that: (i) inputs provided by users to instruct the generation process are not distributed without consent, and if stored, they can be controlled and requested for deletion at any time.
The same applies to the generations, which should also not leak any personal information from the training data.
Overall, all data practices should be transparent and clearly communicated to users prior to generation.

\subsubsection{Access to data}
regulates how organisations handling personal data implement protocols that specify who can access the data, under what conditions, and ensure only qualified personnel with a legitimate need are granted access.
In the context of generative music AI, this requirement applies not only to systems storing data explicitly provided by users during the creative process (such as musical prompts, stylistic preferences, feedback on generated outputs, and uploaded audio/MIDI files), but also extends to all music generated by the system.
This is crucial because generated music, even when intentionally inspired by personal life events, may inadvertently contain or lead to the inference of personal data. The user retains creative control over the prompts and the level of detail they share. For instance, a user might prompt the system with, ``\textit{Write a song about my childhood friend, let's call him ``Jay'', who moved away when we were ten.}''
The AI, following this prompt, could generate lyrics like: \textit{"Jay and I, beneath the summer sky / Built forts of dreams that reached so high / Ten years old, a tearful goodbye / Now just echoes where laughter used to lie."} 
While the user chose to use an alias (``Jay''), the lyrics still contain information about a specific childhood friendship and the emotions associated with it.
Even if the user had provided the real name, the generated lyrics constitute personal data.
The generative system, as the data processor, is responsible for controlling access to these lyrics and any associated metadata (prompts, user ID, timestamps, etc.), regardless of the user's initial intent or level of disclosure.
This is distinct from the model itself; the model generates the content, but the system hosting and providing access to that content must implement the necessary data protection protocols.

Nevertheless, clear access protocols must be designed to facilitate future audits, ensuring a clear and verifiable record of data access and usage, as will be discussed further in the context of Accountability (Facet 7).

\subsubsection{Quality and integrity of data}
used as training material have direct impact on the performance of AI systems and the properties of their generations.
Hence, prior to training, data must be checked for biases, errors, and malicious content.

For music models, having accurate metadata alongside the training material (scores, audio tracks, lyrics, etc.) is a requirement that is often overlooked.
Music metadata is used to consistently identify and describe musical works, their artists, copyright holders, and their associated recordings and scores.
It allows for efficient management and distribution of music, which facilitate search and recommendation \cite{pachet2005knowledge}.
When metadata is accurate, it ensures that artists and copyright holders receive proper credit and compensation \cite{sitonio2018impact}.
Integrity and accuracy of music metadata are thus necessary to acknowledge the contributors of the training musical material, and also enable mechanisms to share potential revenues originating from the generated music.
Finally, in line with previous requirements, we also expect a system to advise if the music model has been trained on music data that can be deemed inappropriate or socially harmful (e.g., offensive lyrics).

\subsubsection{Copyright and licensing} of generated and training data are becoming central in the development of generative systems \cite{sturm2019artificial,Lee2024}, raising complex questions at the intersection of AI and Law.
Copyright grants creators exclusive legal rights to control the reproduction, distribution, and adaptation of their original works, while licensing regulate their use under agreed-upon terms.
These rights are internationally recognised, notably through the Berne Convention of 1886.
The Berne Convention, a cornerstone of international copyright law, establishes fundamental principles such as national treatment (works originating in one signatory country receive the same protection in other signatory countries), automatic protection (copyright exists without requiring formal registration), and independence of protection (protection in the country of origin is independent of protection in other countries).
Generative AI, however, presents novel challenges to these established principles.

By leveraging existing work to automate aspects of the creative process, generative AI challenges conventional definitions of authorship, originality, ownership, influence, sampling, and remixing, and thus complicates existing conceptions of media production \cite{epstein2023art}.
The lack of an explicit legal framework to regulate the use of in-copyright works as data to train generative models and the intellectual property of their outputs, is triggering class actions and distrust.
Recent lawsuits have claimed the former use as illegal, while generations are being disputed as potential infringements of copyright's derivative work right 
\cite{samuelson2023genlegal2}.
Consequently, some creative professionals oppose the use of their work as training data for generative systems \cite{samuelson2023generative}.
This is also motivated by the concern that, while contributing to their success, human works are also likely to compete in the marketplace with their outputs \cite{morreale2021mai,drott2021copyright}.

Since the effectiveness of generative systems relies on vast amounts of human-made musical content, it is thus crucial to protect the intellectual property rights of the creatives who -- explicitly or implicitly, contribute training material.
Meanwhile, detailed licensing agreements are needed to establish the boundaries of how copyrighted material can be used, modified, and distributed for generative AI.
For example, creators may require to be recognised and compensated for their contribution, or even forbid the use of their music for generative purposes through licensing.
Overall, the protection of copyright should strike a balance between safeguarding a creator's right to profit from their work (which encourages further creation) and allowing the development of new creative technologies \cite{samuelson2023generative}.

The legal landscape surrounding copyright and AI is rapidly evolving, taking shape in a patchwork of new legislation.
For instance, in the UK, a public consultation on IP/Copyright related to AI was recently submitted in 2024, and is currently under review \cite{uk2024aicon}.
This consultation specifically seeks to balance the needs of the UK's thriving AI sector and creative industries, addressing challenges such as the use of copyrighted material for AI training.
Another prominent example is ``Tennessee's Ensuring Likeness Voice and Image Security'' (ELVIS) Act, enacted in March 2024 \cite{babl2024elvis}.
The ELVIS Act prohibits the unauthorised use of AI to clone an artist's voice or likeness without consent, making such actions a Class A misdemeanour -- a criminal offence typically punishable by up to a year in jail and/or a fine.
This law builds on existing protections for likeness rights and reflects growing concerns about the misuse of AI-generated content, particularly in the music industry.
The ELVIS Act serves as a legal precedent and aims to influence similar legislation in other jurisdictions.

Despite the ongoing challenge of establishing legal frameworks for generative AI, developers should already attempt to harmonise engineering and copyright requirements \cite{zhong2023acc}.
Through documentation and sound metadata practices, a responsible music system uses material that is either of public domain or whose licensing and terms of use allow for training generative models.
If the training material includes in-copyright data, its use should also comply with local regulations.
For example, the EU has specific exemptions for text and data mining (TDM), allowing non-profit research organisations to freely use copyrighted materials for the purpose of creating new knowledge \cite{geiger2018exception}.
Commercial entities can also engage in TDM, but in this case, copyright holders have the option to \textit{opt out} of this provision.
Other countries, e.g., US, may consider TDM activities as \textit{fair use} of copyright works \cite{samuelson2023generative}.
In the context of generative AI, the concept of fair use aims to strike a balance, allowing for innovation and the creation of new works, while preventing simple reproduction or exploitation of copyrighted material.
However, the application of fair use to generative AI is still a developing area of law, requiring a careful balancing of the rights of copyright holders with the potential for innovation. 

Overall, a generative system should also provide guidance or clear and comprehensive information about the copyright of the generations.
For example, the user terms of AIVA\footnote{\url{https://www.aiva.ai}} only assigns copyright to the user if they pay for their premium plans; otherwise, copyright is fully owned by AIVA.

\subsection{4) Transparency}

This requirement is closely linked with the principle of explainability and encompasses transparency of elements relevant to generative systems, e.g., the data, the system, and the business model.
Hereafter, we outline the main elements influencing the transparency of these systems, pointing the reader to the systematic review by \citeauthor{batlle2023transparency} for an in-depth analysis of the literature in the music domain.

\subsubsection{Traceability} refers to the ability to systematically and comprehensively track and document the various decisions, components, and outputs involved in the development and functioning of a system.
Under this lens, documentation for traceable generative music models and systems should follow established guidelines.
For instance, \textit{Datasheet for Dataset} \cite{gebru2021datasheets} is an approach aimed at standardising dataset documentation within the ML community.
The goal is to encourage the inclusion of a datasheet for every dataset, providing comprehensive details about its characteristics, objectives, and sources.
Instead, the \textit{Model Cards} proposed by \citeauthor{mitchell2019model} are concise documents for providing benchmark evaluations of ML models.
These documents offer insights into the model's performance, intended use cases, and potential biases.
\citeauthor{hupont2023documenting} introduced a use case documentation framework referred to as ``\textit{use case cards}''.
This system, inspired by use case modelling within the Unified Markup Language (UML) standard, stands apart from other methodologies by focusing on the intended purpose and operational use of an AI system.

In the generative music AI field, unfortunately, as reported by \citeauthor{morreale2023data}, still the majority of datasets used for music generation are poorly documented, if documented at all.
In addition, AI-generated music requires robust watermarking mechanisms, both to distinguish it from human-created compositions and to protect against potential harms like deepfakes and misattribution.
This is exemplified by solutions like Meta's \textit{AudioSeal} \cite{roman2024proactive}, Google's \textit{SynthID} \cite{google2023synthID}, and Microsoft's \textit{WavMark} \cite{chen2023wavmark} -- all general-purpose watermarking methods that were also tested on AI music. 

When watermarking is not feasible or has been bypassed, algorithms for AI-music detection can serve as a complementary approach to identify synthetic content.
However, these detection methods face challenges in robustness and generalization, requiring ongoing research and development \cite{afchar2025ai}.

\subsubsection{Explainability} refers to the ability of a generative system to provide appropriate explanations for its outputs.
Prof. Byran-Kinns has provided an in-depth analysis of how explainable AI (XAI) can be applied to the artistic domain \cite{bryan2024exploring}, including music, as documented in the outcomes of the XAIxArts 2023 workshop \cite{bryan2023explainable}.
First, it is important to note the distinction between functional explanations of AI, which centre on technical aspects and accurate outcomes, and the artistic context, where there are no definitive right or wrong answers.
In artistic endeavours, results are often unexpected and unconventional, making it challenging to comprehend how ML models generate human-like outputs.
This presents a distinct challenge for researchers aiming to make AI systems less confounding and more intuitive.
Additionally, there is a pressing need to underscore the significance of real-time explanations and the dynamic nature of XAI in creative environments, where continuous feedback is key.
Ultimately, explanations should be tailored to individual creators and the broader audience, taking into consideration diverse backgrounds and expectations of AI-generated art.

An additional level of explainability is also achieved when systems can trace each generation back to the musical material that most directly inspire it.
This transparency is a key advantage of template-based generative methods \cite{hadjeres2020vector}, which blend pre-existing musical ideas (e.g., patterns, motives, samples) in a combinational settings.
Notably, this approach offers the potential to properly credit the original artists \cite{de2023harmonic}.

\subsubsection{Communication} is key to inform users about the capabilities and limitations of the system.
For interactive systems, users should always be aware they are interacting with an system rather than a human.
Additionally, clear explanations should illuminate the purpose of the AI, the parameters that influence its musical output, and any potential constraints or biases embedded within the model.
To empower informed use across various generative platforms, communication should also address the system's advantages, technical boundaries, and associated risks.
This includes providing instructional resources and disclaimers to promote a responsible and effective interaction with the generative system.

\subsection{5) Diversity, non-discrimination and fairness}
This requirement emphasises the need to promote diversity, non-discrimination, and fairness in generative systems by establishing mechanisms to avoid unfair bias, designing for accessibility, and ensuring fair treatment for all users.

\subsubsection{Avoidance of unfair bias}
is pivotal when generative systems may be affected by historical bias, incomplete data, and poor governance models in the datasets used for training and operation \cite{Srinivasan2021}.
Propagating these biases could result in unintended discrimination and prejudice, exacerbating marginalisation of certain groups \cite{jiang2023ai}. 
It is key to remove identifiable discriminatory bias during the data collection, and oversight processes should be established to analyse and address biases.

In generative music systems, the limitations often arise from the characteristics of the training data, because the model is restricted to producing outputs that align with the established distribution of the input (training) material. 
%
%
This inherent limitation may introduce bias, particularly for underrepresented music cultures. 
While it may be unrealistic to expect all models to fully address this issue, it is crucial to provide a quantification of the type of music data used for training, along with statistics on the training corpus. 
%
%
Nevertheless, the diversification of the datasets used for training cannot be considered as an ultimate goal, but instead only a step towards the ethical and cultural turn that the generative AI music community should confront \cite{Huang2023}.

Under a different lens, the unique capabilities of generative AI lead to the emergence of new aesthetics that could have a lasting impact on art and culture \cite{epstein2023art}. 
As these tools become more prevalent and their use becomes widespread, it remains uncertain how the aesthetics of their outputs will influence artistic creations. 
The accessibility of generative AI could potentially broaden the diversity of artistic outputs by enabling a wider range of creators to engage in artistic practice. 
However, it is also important to consider that the aesthetic and cultural biases embedded in the training data may be reflected and even amplified, potentially reducing diversity.
Additionally, the widespread distribution of AI-generated music may contribute to the creation of future models, establishing a self-referential aesthetic cycle that perpetuates AI-driven cultural norms.

\subsubsection{Accessibility and universal design} is fundamental for allowing people of all ages, genders, abilities, and characteristics to utilise generative systems. 
These systems should not adopt a one-size-fits-all approach and should adhere to design principles that cater to a broad range of users, in compliance with relevant accessibility standards. 
This will facilitate fair access and active involvement of all individuals in both current and future computer-mediated human activities.

Accessibility in generative music AI is often overlooked.
Many systems rely on textual or traditional musical inputs, which may present challenges to certain user groups.
To enhance inclusivity, we must explore alternative input modalities beyond text or music notation.
One promising approach is utilising eye motion as input, potentially followed by haptic feedback for output.
The \textit{EyeHarp} project \cite{Vamvakousis2016} demonstrates this concept, empowering people with severe motor disabilities to learn, perform, and compose music through gaze control.
Furthermore, it is important to distinguish between the model's capabilities and the overall system or interface. 
While a model may have limitations in terms of accessibility, the system can play a crucial role in addressing these gaps. 
This can be done by designing inclusive and diversified interfaces to provide meaningful interactions for users who have difficulty with conventional input methods.
Recognising that some users may face difficulties with a given design, while developing interfaces tailored to specific user categories could help mitigate some of these issues \cite{Agres2021}.
Overall, more research efforts are needed to explore how music human-computer interaction methodologies \cite{holland2019new} can enhance the accessibility of generative music systems.

\subsubsection{Stakeholder Participation} is key to ensure the development of trustworthy AI systems. 
When developing generative AI music systems, it is beneficial to engage and collaborate with creative professionals, ensuring their active involvement in the design, co-creation, and evaluation processes. 
Their level of engagement should be quantifiable throughout the development stages, providing technical insights for system improvement and collecting feedback on the anticipated ethical impact of each solution.

We advocate for a methodological shift, recognising that the perceived needs of artists by AI experts may differ from what they actually require.
This shift should be supported by frameworks guiding the design, development, and deployment of creative systems in ways that are meaningful to musicians \cite{vear2023human}.
Additionally, there is a need to acknowledge and address potential gaps in methodologies, paradigms, systems, and interfaces that are currently overlooked or underrepresented in the field.
The powerful examples of collaborative initiatives in works such as \cite{sturm2019machine} demonstrate the potential of open dialogue between generative AI practitioners and professional musicians.
By embracing stakeholder participation and seeking regular feedback, even post-deployment, we move towards building generative systems that are not only technically innovative but truly responsive to creators' needs.

\subsection{6) Societal and environmental well-being}

This requirement mandates AI development to prioritise fairness, the well-being of society and the environment, with a long-term outlook for the benefit of humanity.

\subsubsection{Sustainable and environmentally friendly AI}
emphasises development and use must prioritise environmentally conscious practices throughout the entire system's lifecycle to minimise its ecological impact.
Overall, generative models requires significant computational power, which translates to high energy consumption and a substantial carbon footprint \cite{chien2023carbon}.
Before the Generative AI outbreak, Google's total electricity consumption was 18.3 TWh in 2021, with AI accounting for 10\%-15\% of this total.
The latter portion was already estimated to reach the same annual energy demand of Ireland (29.3 TWh) \cite{de2023growing}.

As these systems become more sophisticated, so does their energy demand.
The cycle of developing a generative model can be divided into training and inference (or generation).
Since the footprint of a model is directly tied to its parameter count, minimising parameters should be a key objective to achieve efficiency.
For example, audio-based music generative models have different degrees of parameter efficiency:
\textit{Jukebox} \cite{dhariwal2020jukebox} by OpenAI counts 6B parameters and was trained on a grid of 1k+ V100 GPUs for 8 weeks;
Meta's \textit{MusicGen} \cite{copet2024simple} was released in three versions -- 300M (small), 1.5B (medium) and 3.3B (large) parameters, trained on 32, 64 and 96 GPUs, respectively;
and \textit{AudioLDM v2} \cite{liu2023audioldm} counts 1.5B parameters and was trained on 8x A100 GPUs.

To promote transparency and allow for informed decision-making, music models should disclose their parameter count and the hardware, time, and energy resources used during training (\textit{training footprint}).
This also applies to the generations (after a model has been trained), as \citeauthor{chien2023carbon} estimated that for ChatGPT-like services (the systems), inference dominates emissions, in one year producing 25x the carbon-emissions of training GPT-3 (the model).
Therefore, a responsible music generative system should provide indication of the environmental footprint created after generating a whole song or a part of it (\textit{inference footprint}).

\subsubsection{Social impact}
raises awareness on the effects that the use of an AI system could have on social agency, skills, and people's physical and mental wellbeing.
\citeauthor{henrik2023genai} investigated the impacts of generative AI, distinguishing between societal implications (\textit{macro}), impacts on sectors, groups, organisations (\textit{meso}), and individual effects (\textit{micro}).
At the individual level, a primary concern with assistive technologies is the potential for \textit{cognitive atrophy}.
If AI consistently performs mentally demanding tasks, including creative work, we risk losing our own capacity for these functions \cite{saetra2019ghost}.
Just as widespread calculator use led to a decline in mental arithmetic skills, reliance on generative models could diminish our capacity for independent music creation.

Nonetheless, generative music systems can also be designed or used to bring societal benefits.
Models for wellbeing can create or personalise music to improve listener affective state, aiding in relaxation and stress reduction \cite{Williams2020genh}.
In education, generation can support the creation of musical examples, and aid in understanding music theories.
For example, music models can encapsulate particular composition techniques, and be used to generate fragments of illustrative materials that students can explore and challenge \cite{holland2013artificial}.
Music AI systems can also be designed to improve accessibility of creative technologies, allowing them to create and experience music in novel ways \cite{craig2024jess}.

Furthermore, as generative music systems may rely on additional annotations sourced from humans (visible and invisible labour), responsible data collection practices are often overlooked and poorly documented \cite{denton2021genealogy}.
This may also lead to the obfuscation of human labour in ML applications \cite{irani2015difference}.
Therefore, if the model uses any human-made annotation (for training, or evaluation), data collection or crowdsourcing should be designed, conducted, and documented ethically, fairly, and with adequate compensation for annotators \cite{morreale2023data}.

\subsubsection{Society and democracy}
can also be impacted by AI systems, making the measurement of these potential effects a critical part of responsible development.
As human expression is deeply intertwined with art, understanding and guiding AI's influence on creativity also contribute to tracking its broader societal implications in the long term.

While generative systems present potential risks to creative occupations, they may simultaneously enhance productivity in others and facilitate the emergence of new roles.
In the music industry, for instance, automation technologies historically expanded the musician pool, even while altering income distribution patterns \cite{hesmondhalgh2021music}.
\citeauthor{henrik2023genai} suggests we must carefully monitor how these shifts alter power dynamics within the workplace and evaluate whether those changes align with principles of decent work, as outlined in the Sustainable Development Goals (SDG) put forth by the \citeauthor{nations2015transforming}.
While generative AI promises significant advancements in economic growth (SDG 8) and innovation (SDG 9), ensuring its positive and socially sustainable impact is crucial.
Although the development of generative AI models is theoretically accessible, the vast resources required create the risk of growing dependence on large tech companies, potentially undermining a fair and equitable future \cite{widder2023open}.
This also corroborates \citeauthor{morreale2021mai}'s view, arguing that generating realistic music offers negligible advantages to audiences and musicians, while primarily benefiting corporate entities.
In addition, while in principle generative systems can create hundreds of songs per minute and accelerate the creative process through rapid ideation, this might also hinder creative development by removing the initial period of prototyping associated with a ``tabula rasa''\cite{epstein2023art}.

Central to the development of societally responsible generative systems is the provision of mechanisms for: (i) detecting potential cases of plagiarism and IP infringement from AI-generated music; (ii) account for compensation schemes and/or sharing revenues from the generations with the artists that contributed training material.
For the first, \citeauthor{zongyu2021steal} introduced the ``\textit{originality report}'' -- a framework for measuring the extent to which an algorithm copies from the input music on which it is trained.
To enable human authors to continue their socially valuable work and invest time and effort in literary and artistic creations, it is advisable to introduce remuneration rules that offer financial support for human creativity \cite{senftleben2023generative}.
As pointed by \citeauthor{morreale2023data}, one way to implement this is to compensate those who contributed training material, thereby paying copyright holders for their labour and disclosing these amounts whenever possible \cite{denton2021genealogy}.
For generative systems asking for fees or subscription plans, another way is to share revenues and royalties with the artists that contributed training data, as increasingly demanded by artists \cite{clancy2022artist}.
The latter is a form of output-oriented levy system that imposes a general payment obligation on all providers of generative AI systems.
\citeauthor{senftleben2023generative} advocates that output-oriented AI levy systems can be also be combined with mandatory collective rights management.
Collected levies serve two purposes: compensating authors and right holders according to established plans (e.g., the amount of training material contributed); funding social and cultural funds to assist authors disrupted by generative AI.

Directly addressing the unique challenges of the music industry, \citeauthor{jacques2024protecting} propose a model that builds upon the concept of an output-oriented levy system.
In addition to proposing a dual-licensing model comprising two key elements -- (1) licensing requirements for commercially exploited AI-generated content, akin to those for human-produced songs, and (2) licensing AI services that train on copyright-protected material -- they propose an AI-royalty fund specifically tailored for the music industry.
This fund, financed by the proposed licensing fees on AI service providers (calculated based on prompts and outputs created within the UK using IP addresses or a share of subscription fees from UK-based subscribers), would be managed by a collaborative body comprising the Council of Music Makers\footnote{The Council of Music Makers is a UK coalition uniting organisations representing diverse music creators (songwriters, composers, performers, producers, etc.) to collectively advocate for their shared interests within the music industry.} and The Independent Society of Musicians\footnote{The Independent Society of Musicians (ISM) is a UK professional membership organisation that directly supports individual musicians with resources such as legal advice, business guidance, and career development, while also campaigning for their rights.}.
This body would oversee the fund to support human music creators through direct grants, skills development programmes, and targeted aid for genres particularly vulnerable to AI displacement, such as less commercially popular styles, similar to the Music Venue Trust’s model for the live sector\footnote{The Music Venue Trust’s model focuses specifically on supporting and preserving grassroots music venues in the UK—the crucial physical spaces where live music is performed—through fundraising, advocacy, and providing resources to ensure their survival and contribution to the music ecosystem.}.
By ensuring fair remuneration for the use of copyright-protected works in AI training and fostering an inclusive music ecosystem, this model seeks to uphold artistic integrity and economic sustainability within the music industry amidst rapid technological change.

In sum, there are compelling economic and strategic reasons for AI developers to properly credit and compensate artists.
Without fair compensation mechanisms, artists are increasingly turning to data poisoning and other defensive techniques to protect their creative works from unauthorised AI training (see for example HarmonyCloak \cite{meerza2024harmonycloak}, a defensive framework that embeds imperceptible noise into the music to make it ``unlearnable'' by AI models).
Financial sustainability in the arts ecosystem is crucial for AI development itself -- if fewer people can afford to be artists full-time due to lack of compensation, this directly reduces the high-quality training data available for future AI systems.
Finally, creative professionals represent a key customer segment for AI tools.
They are much more likely to adopt technologies they know are ethically developed and free of copyright infringement, especially since many are contractually liable for any infringements in their professional work.
These incentives align business interests with ethical practices, creating a foundation for sustainable AI development that respects creative labour.

\subsection{7) Accountability}

This requirement corroborates all the others by ensuring that AI systems are accountable for their responsible design, implementation, and impact throughout their lifecycle.

A proprietary generative system should ensure \textbf{auditability}, allowing internal and external auditors access (upon request) to assess its compliance with the responsible features defined before.
Audit reports must be publicly available.
Prior to their development, potential \textbf{negative impacts} should be identified, assessed, minimised and openly communicated, while conducting risk assessments with the involved stakeholders and protecting those who may raise concerns.
This particularly applies when a generative system does not accommodate one or more responsible feature, and can be facilitated by red teaming methods \cite{feffer2024red} or via questionnaires, e.g., the Algorithmic Impact Assessment (AIA) \cite{metcalf2021aia}.

Auditability may come from internal assessments or, increasingly, from external certification bodies that provide independent validation of responsible AI practices.
These certifications can serve as a readily accessible form of audit report, promoting transparency and accountability within the generative AI ecosystem.
An example is Fairly Trained, which also intersects the ``Access to Data'' requirements in Facet 3 (Privacy and Data Governance).
Fairly Trained\footnote{\url{https://www.fairlytrained.org}} is a non-profit organisation certifying generative AI companies that prioritise creator consent and fair compensation.
Their Licensed Model (L) certification distinguishes AI models trained solely on licensed data, guaranteeing artists' rights are respected.
This certification is available for entire companies, specific products, or individual models, with costs varying based on revenue.
Within music AI, several key companies and projects have obtained Fairly Trained certification.
For instance, \textit{Endel}, which creates AI-powered soundscapes for wellbeing, and \textit{LifeScore}, known for AI-generated remixes that preserve artists' musical style, are both certified.
Additionally, specific products like \textit{Boomy}, a generative music platform enabling users to create and monetise original songs, and models such as Jen's audio models, used in their ethically trained text-to-music platform, have also received Fairly Trained certification.

Implementing responsible features also comes with the definition of \textbf{trade-offs}, as requirements may inherently conflict in the design of a music model or system.
For instance, a model designed for highly realistic music generation (high accuracy) might lack explainability and have significant computational costs, harming environmental sustainability.
Driven by the target use of the generative model/system, trade-offs should be carefully selected to prioritise the minimisation of risks to ethical principles.
If no ethical trade-offs can be found, the development or use of the model should be reconsidered, and the decision-makers should be held accountable for the potential impacts.

Finally, a generative system should provide \textit{redress} mechanisms whenever it fails to meet responsible expectations and unjust adverse impact occurs (e.g., generations contain offensive content, the model excessively plagiarise the training material).
These usually come in the form of compensation, direct correction through feedback, legal support, etc.

\section{Opportunities and Directions}\label{sec:discussion}

The responsible features presented in this work (c.f. Table~\ref{tab:features-all-c} in the Appendix), serve as a preliminary foundation to stimulate an interdisciplinary dialogue aimed at identifying the core requirements for Responsible AI in music.
Below, we discuss their potential to drive innovation through trustworthiness, anticipate future transformations, and foster interdisciplinary collaboration for their implementation.

\subsection{Trustworthiness as a driver for innovation}

Ethical considerations are sometimes perceived as constraints on technological progress \cite{floridi2021establishing}.
However, here we posit that trustworthiness, rather than hindering innovation, serves as a powerful catalyst for creativity and advancement in generative music AI.
By embedding responsible features into the design process, we shift the paradigm from viewing ethics as an afterthought to recognising it as a core design principle \cite{van2007ict}. 

The features provide developers with a responsible creative space to design generative music models and systems that are \textbf{ethical-by-design}.
Each feature encapsulates a research direction per se, with its own peculiarities and challenges.
Features such as ``\textit{Creative Feedback (HITL)}'' (feature 2, alias F2), ``\textit{Controllability (conditioning)}'' (F3), and ``\textit{Generation Explainability}'' (F25) are not merely ethical safeguards; they are also research avenues that push the boundaries of human-AI interaction.
For example, implementing ``\textit{Creative Feedback (HITL)}'' necessitates the development of sophisticated mechanisms for capturing and interpreting nuanced user input, driving research in interactive machine learning and multimodal interfaces \cite{tanaka2020multi}.
Similarly, ``\textit{Generation Explainability}'' requires advancements in interpretable AI, fostering a deeper understanding of the creative processes of these systems and enabling more meaningful collaborations between humans and AI \cite{bryan2024exploring}.
These technical challenges, inherent in the pursuit of trustworthiness, thus reframe existing research problems in music generation, giving them new significance and urgency.
They shift the focus from solely pursuing musically realistic outputs -- often a primary goal in evaluating generative music AI models (e.g., \textit{Model Evaluation}'' (F10), \textit{Music Evaluation}'' (F11)) -- towards a broader landscape of responsible and human-centered design.

Trustworthiness is also central for user \textbf{acceptance} and \textbf{adoption}.
While dedicated research on generative music AI is lacking, studies on AI-based image tools \cite{xu2023everyone,palani2022don} provide preliminary evidence that factors like perceived usefulness, trust, and ease of integration are also relevant for adoption in creative practices, suggesting similar principles likely apply \cite{kelly2023factors}.
Artists and musicians are increasingly cautious of technologies that could potentially devalue their skills or infringe upon their rights as demonstrated by initiatives like the ``\textit{Human Artistry Campaign}'' and ``\textit{AI for Music}'' (see Introduction).
The framework's emphasis on features like ``\textit{Prompt Leakage Prevention}'' (F16), ``\textit{Training Metadata Integrity}'' (F17), ``\textit{Generative Reuse of Music Data}'' (F20), and ``\textit{IP Validation}'' (F40) directly addresses concerns surrounding data privacy, copyright, and plagiarism.
By adhering to these responsible features, developers can cultivate \textbf{trust} with users, encouraging them to accept these tools \cite{jacovi2021formalizing}.
This aligns with the growing movement towards ethical AI, exemplified by initiatives like the ``Fairly Trained'' certification, which recognises models trained solely on licensed data.
If the evaluation of trustworthiness is intuitive and transparent, this also opens up a market where responsible development is not only recognised but also rewarded, potentially through certifications, ratings, or preferential treatment in music platforms.

\subsection{Anticipating transformations: a proactive stance}

The rapid evolution of generative AI in creative domains necessitates a proactive approach to ethical considerations.
We cannot simply react to ethical challenges as they arise; instead, we should anticipate future transformations and their potential implications to shape the development and deployment of these technologies responsibly \cite{sarewitz2011anticipatory}.
For example, in fields like Human-Robot Interaction (HRI), researchers have long recognised the importance of anticipating challenges associated with integrating robots into human environments \cite{sparrow2006hands,goodrich2008human}.
By studying human-robot interactions and envisioning future scenarios -- even those that seem far-fetched at present -- their work involves anticipating potential issues long before the technology becomes commonplace.
This has already contributed to the design of robots that are not only functional but also socially acceptable, safe, and ethically sound.

Similarly, we can anticipate the future developments and use cases of generative music AI.
This means engaging in speculative yet grounded thinking about how the technology might evolve and the impacts it could have on music creation, consumption, and the broader cultural landscape.
For instance, which new applications will generative music AI enable that we have not even thought of yet, and how will they redefine our relationship with music as artists or listeners? Which challenge will streaming platforms face when they are inundated with AI-generated music, and how might this affect discoverability for human artists? Could AI systems curate personalised soundtracks adapting to our moods, activities, and even physiological states in real-time, and what are the implications for privacy and autonomy in such a scenario? How might our understanding of authorship and originality evolve when AI can generate infinite variations of a musical theme, potentially blurring the lines between inspiration, imitation, and plagiarism? Moreover, what new forms of musical expression might emerge from human-AI collaboration that are impossible for either humans or AI to achieve alone? Will we see the rise of entirely new genres or performance practices or their latent re-amplification?
These are examples of speculative questions we can be asking now to anticipate new ethical implications, even if the technology to fully realise these scenarios is not yet mature.

Our preliminary set of responsible features offers an initial roadmap for navigating this uncharted territory.
While not yet exhaustive, these features provide a structured approach to start considering the ethical and societal implications raised by the questions above as the technology continues to evolve.
For instance, features related to \textit{Societal Impact}, such as ``\textit{Responsible Data Collection}'' (F38) and ``\textit{Social Purpose}'' (F39), prompt us to consider how AI-generated music might affect the livelihoods of human musicians and the accessibility of music creation.
Features under \textit{Society and Democracy}, including ``\textit{IP Validation}'' (F40) and ``\textit{Revenue Sharing}'' (F41), encourage us to examine the potential for AI to disrupt existing music ecosystems, challenging our notions of copyright, ownership, fair compensation, and the very definition of musical authorship in the platform era.
Moreover, features under the umbrella of Accountability (Pillar 7), like ``\textit{Audit Access}'' (F42), ``\textit{Minimisation of Negative Impacts}'' (F43), and ``\textit{Generative Redress}'' (F45) push us to plan for potential harms and establish mechanisms for redress when things go wrong.

An example of this anticipatory approach is exemplified by the ``\textit{Red Button}'' (F4) feature -- the ability to immediately halt an AI system's music generation if it becomes unpleasant, disturbing, or contains offensive content.
While seemingly futuristic, this feature encourages us to anticipate scenarios where real-time control over AI-generated music might be necessary, such as in public spaces or interactive performances.
Early intervention, guided by an anticipatory framework, can thus allow us to shape the trajectory of generative music AI, maximising its benefits while minimising harms from their inception.
This is already crucial in areas like copyright and intellectual property, where early decisions about data usage and licensing, as emphasised in features F19-F21, can have long-lasting consequences.
Similarly, proactively engaging with features like ``\textit{Revenue Sharing}'' (F41) allows us to explore new economic models that could ensure fair compensation for artists whose work contributes to the training of AI models.

While the implementation of such features might present technical and logistical challenges today, beginning the discussion and developing the groundwork now is vital for creating a sustainable and equitable future for music in the age of AI.
This involves not only anticipating potential problems but also envisioning positive scenarios where generative AI empowers new forms of creativity, expands access to musical creation, heritage, and enriches the musical landscape.

\subsection{Interdisciplinary collaboration: bridging the gaps}

The development of responsible generative music AI cannot be achieved through isolated efforts.
It demands a concerted, interdisciplinary approach that bridges the gaps between various stakeholders, each bringing their unique expertise and perspectives to the table \cite{kaila2023ethically}.
This work underscores the importance of such collaboration as a prerequisite for building a robust framework for the responsible design of these technologies.

The responsible features presented here serve as a catalyst for interdisciplinary dialogue by establishing \textbf{common ground} -- a shared space, a vocabulary and a set of concrete objectives -- around which diverse stakeholders can align.
This group must include AI researchers and developers, music professionals (composers, performers, producers), ethicists, legal experts, music industry stakeholders (labels, publishers, streaming platforms), policymakers, and end-users (listeners, amateur musicians).
Their roles are interconnected and essential: AI experts implement the technical solutions, providing insights into feasibility and limitations; musicians offer creative guidance, evaluate musical outputs, and articulate their needs, rights and concerns regarding AI's role in the future of music; ethicists help ensure alignment with human values, societal well-being, and fairness principles; legal experts navigate the complex and evolving landscape of copyright, intellectual property, and data privacy, which varies significantly across countries; industry stakeholders shape economic models, distribution strategies, and user adoption patterns; policymakers create regulatory frameworks and incentive structures that promote responsible innovation; and end-users provide feedback on their experiences, preferences, and concerns, ultimately shaping the demand and acceptance of the technology.

While to date efforts have often been isolated within these respective communities, the rapid advancement of Generative AI is increasingly necessitating a convergence of perspectives \cite{bommasani2021opportunities}.
However, enabling this exchange is not without its challenges.
Communication barriers often arise due to the specialised language, technical concepts, and methodologies inherent to each discipline.
For example, a computer scientist might focus on algorithmic efficiency and model accuracy, while a musician might prioritise expressive control and the originality of the generated music.
Bridging this gap requires a concerted effort to demystify technical jargon, establish common definitions, and cultivate a shared understanding of core concepts.
Differing priorities and potential power imbalances between stakeholders can also lead to conflicts.
For instance, industry stakeholders might prioritise commercial viability and market dominance, potentially at the expense of artist compensation or the diversity of musical output.
Policymakers might struggle to balance innovation with regulation, potentially stifling creativity or failing to adequately address ethical concerns.
Addressing these challenges requires open dialogue, transparent decision-making, and a commitment to finding solutions that balance the interests of all parties involved.
Moreover, there is a need for more established forums and initiatives that facilitate sustained dialogue and collaboration across these diverse groups. 
Projects like ``Music RAI'' and ``AI:OK'' are making important strides in this direction, bringing together researchers, musicians, and industry stakeholders to address specific issues like \textit{bias} and the \textit{impact} of AI on the music ecosystem.

This work aims to bridging these gaps by grounding the discussion in concrete, actionable features.
By focusing on specific design considerations, a more productive and inclusive dialogue can be fostered.
The first step towards enabling this is to establish a \textbf{common terminology} and understanding of basic technical concepts, creating a shared foundation for communication.
For example, AI researchers should make an effort to explain technical details and their implications in ways that are accessible to non-experts, while musicians and other stakeholders should be open to learning about the current capabilities and limitations of AI systems.
This requires promoting \textbf{interdisciplinary education} and \textbf{outreach}.
AI researchers would need to share some understanding of the history and cultural significance of different musical traditions, copyright law, and the nuances of the music industry's economic and social structures.
Conversely, musicians, music professionals, and other stakeholders must be familiarised with the basic principles of AI, including its strengths, weaknesses, and potential societal impacts.
This can be done via workshops and joint-research initiatives.

The approach for collaboration should be multifaceted.
Firstly, it should be iterative and cross-disciplinary, where features are not designed in isolation but are co-created and refined through a continuous feedback loop involving all stakeholders.
This ensures that complex requirements -- those involving technical, legal, ethical, and economic considerations -- are adequately addressed.
For instance, a feature like ``\textit{Revenue Sharing}'' (F41) requires input from legal experts to define fair licensing models, from industry stakeholders to design viable economic mechanisms, from ethicists to ensure equitable distribution of benefits, and from AI researchers to develop the technical infrastructure for tracking and distributing royalties.
Secondly, the approach must also accommodate domain-specific requirements that primarily pertain to a particular community.
For example, features related to ``\textit{Model Evaluation}'' (F10) might be primarily driven by AI researchers, while features related to ``\textit{Artist Involvement}'' (F7) would involve musicians and creative professionals.
This dual approach -- combining cross-disciplinary collaboration with specialised ad-expertise -- ensures that the resulting framework is both comprehensive and nuanced, addressing both general and domain-specific needs.
It is worth highlighting that some features may not initially be valued by all the actors involved.
For example, music artists may not perceive the immediate relevance of detailed ``\textit{Model Evaluation}'' metrics (F10).
However, it is the \textbf{union} of all these perspectives, and their \textbf{iterative}, \textbf{cross-domain contamination}, that yields more complex, nuanced, and ultimately robust requirements.

\begin{figure*}[t]
    \centering
    \includegraphics[width=.7\linewidth]{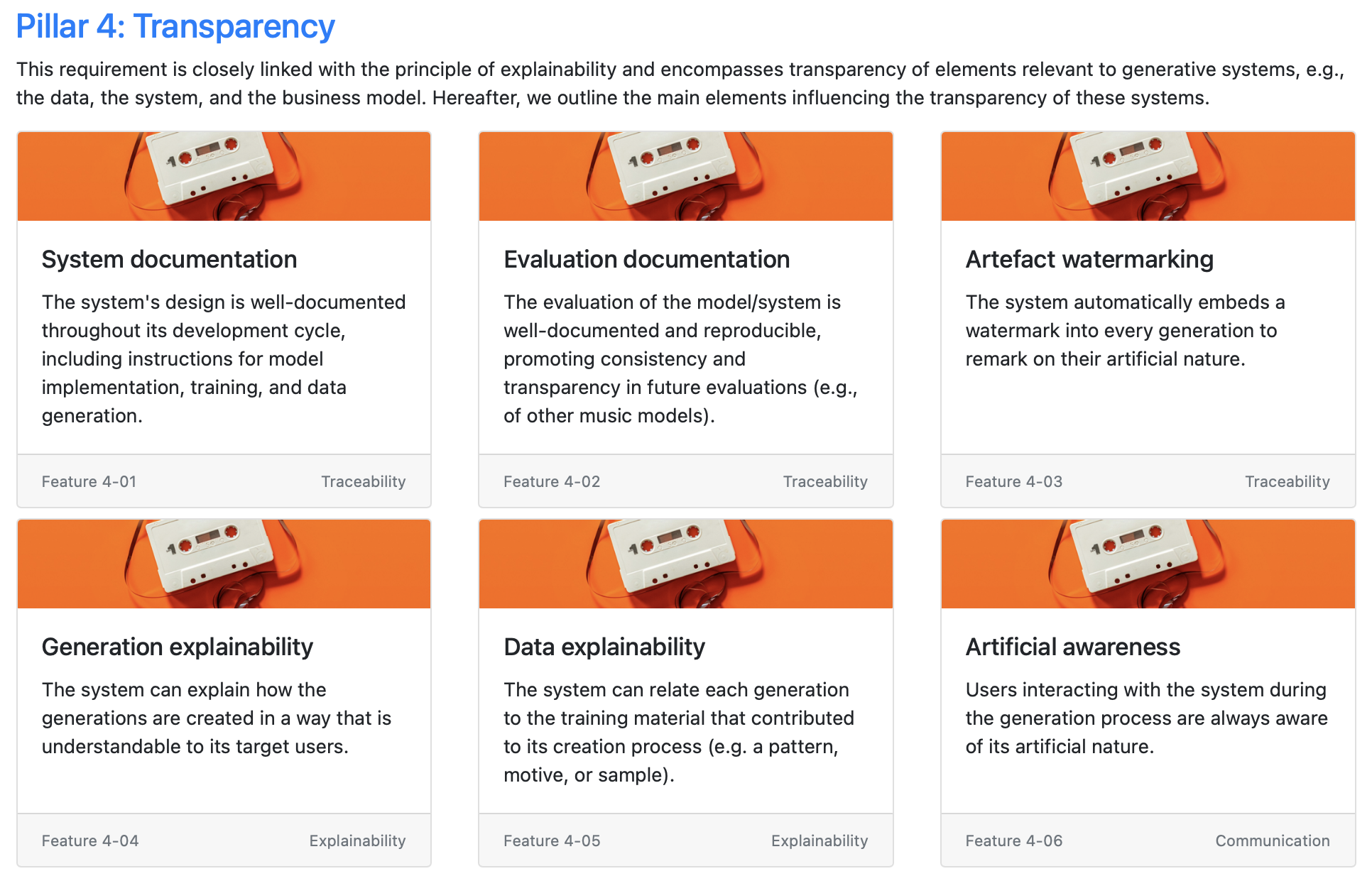} 
    \caption{Screenshot of the companion website showcasing the 45 responsible features for generative  AI music. The website presents each feature in an individual card format, simplifying navigation and promoting wider understanding of the framework.}
    \label{fig:raim-website-screen}
\end{figure*}

\section{Operationalising the Responsible Features: \\a tentative roadmap for framework design}

While opening a discussion on the responsible features is a preliminary step, translating these principles into a functional framework requires establishing a more comprehensive and well-articulated set of measurable requirements, clear mechanisms for their evaluation and documentation, and a shared vision for their iterative refinement.
This necessitates a multi-faceted approach driven by stakeholder engagement and a realistic appraisal of the inherent challenges.

\subsection{Stakeholder evaluation and refinement of features}
A critical first step in implementing the framework is a comprehensive evaluation of the proposed features by the relevant stakeholders.
This involves soliciting feedback from AI researchers, music professionals, ethicists, legal experts, industry stakeholders, and end-users to assess the perceived importance, relevance, and feasibility of each feature from their respective perspectives.
This process will employ a mixed-methods approach, incorporating surveys, structured interviews, focus group discussions, and Delphi studies to gather both quantitative and qualitative data.
The collected insights will be used to refine the feature set.

The objective of this evaluation is threefold: (1) to rank the features based on their perceived importance and urgency, allowing for prioritisation during implementation; (2) to identify potential gaps, ambiguities, or limitations in the current feature set, paving the way for refinement and expansion; and (3) to foster a sense of ownership and shared responsibility among stakeholders, ensuring that the framework reflects a broad consensus and addresses the diverse needs of the community.
The insights gained from this process will inform subsequent iterations of the framework, ensuring it remains relevant, comprehensive, and responsive to the evolving landscape of generative music AI.
Versioning mechanisms will be put into place to ensure consistency while allowing for referencing older versions of the framework/features and tracking their evolution.

\subsection{Framework implementation}
Once a first stable version of the feature set is finalised, we advocate for the implementation of an actionable framework that enables the translation of these principles into concrete design choices and operational practices.
\textit{How may a responsible framework for generative music AI look like?}

First of all, it is crucial to acknowledge that not all features may be equally applicable or achievable for every music model/system.
The application of the framework should be flexible enough to account for the specific context, objectives, and constraints of each generative method.
Developers are expected to engage in a thoughtful and transparent process of selecting, prioritising, and implementing features based on their relevance to the solution being developed.
A clear articulation of these trade-offs, providing a justification for the decisions made at the onset of the project already aligns with the principles of ``Accountability'' (Pillar 7).
The primary goal is not to ``tick all the boxes'' -- but to recognise that those features are relevant, and to document the model/system to inform potential users on where the contribution lies in the responsible landscape.
This puts the emphasis on ethics by design \cite{van2020embedding} while facilitating users -- ranging from public stakeholders to artists and industry professionals, to find the most suitable and ethically compliant music AI solution to reuse, given their \textit{requirements} and \textit{constraints}.
These might include internal organisational policies, regulatory guidelines, environmental compliance, or specific technical infrastructure requirements.
For example, a company might have policies regarding data provenance, intellectual property rights, or compatibility with existing software ecosystems that would influence their choice of a music AI solution.
Similarly, a user might be bound by legal requirements related to accessibility, or ethical guidelines established by a professional body.

Therefore, the framework is not intended as a rigid checklist, but rather as a set of well-defined and actionable guiding principles.
It provides a structured, yet flexible, approach to responsible development.
To this aim, we anticipate the following questions to drive its implementation:
(Q1) How can we measure a music model/system's adherence to each responsible feature?
(Q2) Who should be involved in this evaluation process?
(Q3) How can we ensure transparency and open access to the evaluation documentation?

\subsubsection{Q1: Evaluation.}
An effective framework should have the capacity to quantitatively and qualitatively assess a music model or system's adherence to the defined responsible features.
This necessitates a clear articulation of evaluation strategies -- for each feature -- starting from their initial design and proposition to ensure their actionability.

A systematic, interdisciplinary approach that harmonises the current methodologies and practises in the respective fields is crucial for developing and reusing robust evaluation metrics.
This should involve combining quantitative measures, such as computational metrics for detecting music leakage or user ratings of explainability, with qualitative assessments like expert evaluations and user studies.
For example, the evaluation of feature ``\textit{Generation Explainability}'' (F25) could encompass user ratings quantifying the understandability of a system's explanations, alongside objective measures assessing the fidelity of these explanations to the actual generative process.
Similarly, for ``\textit{Accessibility and Universal Design}'' (features 32-34), quantitative data on usage patterns across diverse user groups could be integrated with qualitative feedback from accessibility experts and users.
Assessing the ``\textit{Controllability}'' (F3) feature might involve measuring the range of musical parameters users can manipulate and the granularity of control they have.
This could be quantified by counting the number of controllable parameters (e.g., tempo, key, instrumentation) and the number of discrete values each parameter can take.
User studies could then assess how effectively users can achieve their desired musical outcomes using these controls, providing a qualitative measure of controllability's success.
Similarly, for ``\textit{Music Leakage}'' (F8), evaluation could involve statistical comparisons between generated music and the training dataset via similarity measures \cite{velardo2016symbolic}; or measure the ability of a model to reconstruct a full piece starting from a given segment.
Thresholds for acceptable leakage levels would need to be established, potentially involving legal experts to determine what constitutes copyright infringement under different legislations.

Nevertheless, while a comprehensive evaluation of these features remain a challenge that goes in parallel with their formulation, a music model/system can still be documented against the framework to acknowledge its expected coverage of these features.
Acknowledging the feature set is indeed a first step.
To facilitate this multifaceted evaluation effort, the framework should be accompanied by a suite of tools and resources.
These may include: detailed guidelines and supporting documentation for each feature, elaborating on its rationale, potential implementation strategies, specific evaluation protocols, and associated metrics; illustrative examples of already evaluated generative music models and systems, complete with their documentation, to serve as benchmarks; and templates for documenting design decisions, trade-offs, and evaluation results consistently.

\subsubsection{Q2: Interactions.}
The implementation of the framework also requires deciding who will be responsible for evaluating its responsible features.
Self-assessment by AI developers may provide a starting point.
However, inherent biases and the potential for overlooking critical flaws may necessitate a more diverse and independent evaluation process.
A multi-tiered approach, combining self-assessment, community-based crowdsourcing, and independent third-party audits, offers a promising solution.

The process ideally begins with self-assessment by the creators, using the provided guidelines and metrics to evaluate their own model/system against each relevant feature.
This step encourages developers to internalise the principles of responsible AI and proactively identify potential shortcomings.
To complement this evaluation, the framework can promote community-based crowdsourcing on a web platform.
This platform would allow a broader range of stakeholders, including musicians, musicologists, ethicists, legal experts, and end-users, to directly engage in the evaluation process.
A crowdsourced approach would leverage the collective intelligence and diverse perspectives of the community \cite{brabham2008crowdsourcing} to identify issues that might be missed by the developers alone.
The approach draws inspiration from successful examples like Wikidata\footnote{\url{https://www.wikidata.org/}} -- a free, open, and collaborative knowledge base that anyone can edit, much like Wikipedia, but instead of articles, it stores structured data that can be used by computers and humans alike.

\textit{What happens if the crowd disagrees?}
Community contributions and discussions are key for maintaining data quality and resolving disagreements in the crowd \cite{kittur2007he}.
Similarly, a platform may incorporate mechanisms for structured feedback, ratings, and discussions, allowing for a nuanced and transparent evaluation process.
Disagreements can be handled through a structured discussion, similar to Wikipedia's ``Talk'' pages or Wikidata's discussion forums, where users can present arguments, provide evidence, and work towards a consensus.
Oversight by expert moderators, potentially drawn from the stakeholder groups mentioned earlier (ethicists, legal experts, etc.), could help resolve more complex or contentious issues.

Even community-based evaluation may not be able assess all features, especially those related to internal system workings or training data that are not publicly accessible.
Some features, such as those related to data privacy or the specifics of training datasets, may be inherently ``unobservable'' to external evaluators. 
This is where independent third-party audits become essential (F42, ``\textit{Audit access}'').
These audits, conducted by accredited organisations or experts, provide an objective and in-depth assessment of the system's compliance with responsible features.
This approach mirrors initiatives like the "Fairly Trained" certification, which verifies that AI models are trained solely on licensed data.
Hybrid approaches that, while supporting developers' self-assessment, also provides these audits are a good trade-off and would potentially harmonise with the ongoing efforts. 

In sum, the combination of self-assessment, community crowdsourcing, and independent audits could create a robust and multi-faceted evaluation ecosystem.
This ensures that AI music systems are scrutinised from multiple perspectives.

\subsubsection{Q3: Documentation.}

Having established how responsible features will be evaluated (Q1) and who will be involved in the evaluation process (Q2), a robust and transparent system for documenting these evaluations is necessary.
This documentation serves not merely as a passive record, but as an active component of the framework, enabling scrutiny, facilitating knowledge sharing, and ultimately fostering trust and accountability.
The documentation process should capture not only the results of evaluations against each responsible feature, but also the methodologies employed, and any identified disagreement encountered in the previous step.
This creates a living record that supports continuous improvement and informed decision-making by all stakeholders.

To ensure that this documentation is both structured and readily accessible, semantic annotation techniques can be leveraged.
This allows algorithms, their inputs, outputs, training materials, and evaluation outcomes to be described in a machine-readable format.
This involves reusing and extending existing schemas to map a generative music AI and describe the outcomes of the evaluation process.
Ontologies such as the Software Ontology \cite{oberle2009ontology} and the Music Algorithm Ontology in \cite{de2023polifonia} already provide a foundation.
Specifically, the latter offers a schema for describing various aspects of music algorithms, including their type, input data formats, output representations, and intended use cases.
By adopting and extending these ontologies, we can create a semantically rich ``Web of Music Data'' around the framework that connects music models, systems, datasets (with their provenance, licensing, and copyright information), evaluation results, and the responsible features themselves.

This structured, machine-readable documentation fuels the creation of a knowledge base, which in turn can power interactive dashboards and search interfaces -- thereby supporting both Q1 and Q2.
These interfaces allow users -- from researchers and developers to musicians and end-users -- to easily find, compare, and select generative music AI tools based on their specific needs and ethical priorities.
For example, a user concerned with artist compensation might prioritise systems demonstrating strong adherence to ``\textit{Revenue Sharing}'' (F41) and ``\textit{IP Validation}'' (F40).
Another user, focused on understanding the creative process, might prioritise ``\textit{Explainability}'' (features 25-26).
The dashboard, operating on the knowledge base, would provide a transparent overview of each system's performance against the framework, empowering users to make informed choices.

Notably, this approach aligns with and extends broader efforts to improve documentation in the AI field.
``Datasheets for Datasets'' \cite{gebru2021datasheets} and ``Model Cards'' \cite{mitchell2019model} have advocated for standardised documentation practices.
More recently, the ``Croissant'' standard \cite{akhtar2024croissant}, a metadata format for describing datasets, offers a lightweight and machine-readable approach, and its RAI extension introduces a  structured vocabulary for addressing responsible AI specifications \cite{jain2024standardized}.
The proposed framework builds upon these initiatives, adapting them to the unique challenges and opportunities of generative music AI.
The knowledge base approach goes beyond simple documentation, fostering discoverability, enabling complex queries, and facilitating comparisons across systems, thereby promoting a more transparent and accountable ecosystem.

\subsection{Implementation challenges}

We anticipate that the implementation and adoption of the framework will require careful consideration of several factors.
These encompass technical aspects, the nuances of subjective interpretation, incentive mechanisms, and adaptation to the field's dynamic nature.

\subsubsection{Inherent Subjectivity}
Despite efforts to ground feature evaluation in measurable criteria, the inherent subjectivity of music and creativity presents a persistent challenge. As noted in the discussion of Accuracy (Pillar 2), evaluation in music generation is notoriously complex, lacking the standardised benchmarks found in other fields like natural language processing \cite{gehrmann2022gemv2}.
This challenge extends beyond purely technical aspects; many responsible features, such as those related to aesthetic quality, originality (as implicated in features like F20, F26 and F40), and even the effectiveness of ``\textit{Artist Involvement}'' (F7), rely on inherently subjective judgments.
What one listener perceives as original, another may find derivative.
What constitutes ``meaningful'' artist involvement is open to interpretation.

While quantitative metrics can be applied to some aspects (e.g., measuring stylistic similarity to training data to assess originality \cite{sturm2016music}), these may not fully capture the nuances of human aesthetic experience.
This necessitates a multi-faceted evaluation approach, incorporating both quantitative data and qualitative assessments from diverse stakeholders, as discussed in Q2.
Furthermore, the evaluation process itself must be transparent and well-documented, acknowledging the inherent subjectivity and striving for a consensus-based approach, potentially leveraging the community-driven mechanisms outlined in Q3.
The subjectivity challenge underscores the need for ongoing dialogue and iterative refinement of the framework, recognising that differing perspectives and evolving societal values will necessitate adjustments in how features are prioritised and measured \cite{mittelstadt2019principles}.

\subsubsection{Lack of Incentives}
Beyond the technical challenges of measurement and auditing, a fundamental obstacle lies in aligning market incentives with responsible development.
Currently, market forces predominantly reward performance, cost-effectiveness, and rapid innovation, often overshadowing ethical considerations \cite{morley2020ethical}.
This creates a ``give and take'' scenario where developers may be reluctant to invest the additional resources and effort required to implement responsible features if it does not translate into a competitive advantage.

Addressing this requires action on multiple levels.
Firstly, fostering greater awareness and demand for responsible music AI among developers, users, industry stakeholders, and the broader public is crucial.
Initiatives like the Human Artistry Campaign and the ``Principles for Music Creation with AI'' represent steps in this direction, but widespread education and advocacy are needed to shift consumer preferences and industry norms.
Secondly, certification and rating systems, exemplified by ``Fairly Trained'', can play a central role in recognising and rewarding responsible development, providing a market-based incentive for companies to prioritise ethical considerations.

Thirdly, funding bodies and research institutions have a crucial role to play.
By prioritising responsible AI development in the creative domain, they can incentivise researchers and developers to adopt this framework.
One potential approach is to adapt principles from the FAIR data movement (Findable, Accessible, Interoperable, and Reusable) \cite{wilkinson2016fair} to the realm of generative music AI.
A "Musically FAIR" framework, for instance, could establish criteria for responsible AI development, analogous to the FAIR principles for data management. 
Projects funded by these bodies could be required to adhere to the framework, either in its entirety or to a prioritised subset of features deemed essential for minimum levels of acceptance.
This would not only promote responsible development but also create a ``star system'' or tiered certification, where music AI systems are publicly recognised (e.g., through badges or certificates) based on their level of adherence to the framework. Overall, these parallel efforts would creates a clear, market-visible signal of ethical commitment.

\subsubsection{An Evolving Landscape}

The rapid pace of advancements in generative AI necessitates a continuous process of adaptation and refinement of the framework, its associated metrics, and implementation strategies.
The capabilities of these technologies are constantly expanding, leading to new challenges and ethical risks that may not have been foreseen during the framework's initial development.
As noted by \citeauthor{cath2018governing}, governing AI is indeed a dynamic process.
This requires a robust versioning system for both the framework itself and the evaluated models/systems.
As models are updated, their responsible AI assessments must also be revisited and updated, reflecting any changes in their behaviour, training data, or intended use.

What constitutes best practice today may become outdated tomorrow.
Addressing this dynamic landscape necessitates establishing mechanisms for ongoing monitoring, evaluation, and revision, as highlighted in the context of addressing complex societal challenges \cite{kuhlmann2014challenge}.
This could involve creating a dedicated body or working group, similar to standards bodies in other domains, responsible for maintaining and updating the framework.
This group would solicit feedback from stakeholders, track developments in the field, and propose revisions to the framework as needed.
It would also foster a culture of continuous learning and adaptation among developers, encouraging them to stay abreast of the latest research and best practices in responsible AI, potentially through workshops, training programs, and shared resources. Furthermore, this dedicated body can also assess the need of introducing new responsible features to the framework, as they emerge from new technological developments.

\subsection{Iterative Refinement and Future Directions}

The framework is not intended as a static endpoint, but rather as a foundational element for an ongoing, collaborative, and multidisciplinary discussion. Its creation is envisioned as a living process, subject to continuous refinement and expansion. This evolution will ideally be driven by stakeholder feedback, empirical evidence gathered from the framework's application, and advancements in the rapidly developing field of generative AI.
We anticipate that this iterative process could involve various efforts, including:

\begin{itemize}
    \item \textbf{Establishing and Curating an Open Repository.} A publicly accessible, online repository could serve as a central hub for the framework, its associated metrics, detailed implementation guidelines, and illustrative case studies. Such a repository could be designed to encourage community contributions and feedback, fostering a collaborative environment for ongoing improvement. Furthermore, it could map the existing landscape of generative music AI models and systems, annotating them according to their adherence to the framework's features. This would empower users to search for and identify generative tools that align with their specific needs, requirements, and ethical expectations (see Q3). The accompanying website (\RAIMw) represents an initial step in this direction, offering a platform for dissemination and feedback collection for the iterative refinement of the responsible features (see Figure~\ref{fig:raim-website-screen}).

    \item \textbf{Promoting Research and Dissemination.} The broader research community can play a crucial role in advancing the framework through various initiatives. Workshops and discussion groups could facilitate dialogue among stakeholders, ensuring the framework remains relevant and responsive to the community's evolving needs. These events could also serve as platforms for showcasing best practices, sharing lessons learned, and fostering collaboration between researchers, developers, musicians, legal experts, and ethicists. Furthermore, the research community could focus on the development of new evaluation techniques and technical solutions for implementing responsible features. This could involve establishing dedicated tracks or challenges within existing music information retrieval (MIR) and AI conferences, with a specific focus on responsible design.  Key examples include MIREX (the Music Information Retrieval Evaluation eXchange), which runs annual comparative evaluations of MIR algorithms, and ISMIR (the International Society for Music Information Retrieval Conference), the leading conference in the field. These challenges could target specific features, encouraging the creation of, for example, robust metrics for music plagiarism detection or new mechanisms and benchmarks to measure data leakage.

    \item \textbf{Monitoring Impact and Gathering Evidence.} The framework's impact on the development and use of generative music AI systems should be continuously monitored. A dedicated community could emerge to assess its effectiveness in promoting trustworthiness, mitigating potential risks, and fostering a more equitable and sustainable music ecosystem. This could encompass the collection of both quantitative data on framework adoption and usage, and qualitative data, such as success stories, interviews with creatives, and case studies, to understand its practical implications. Such ongoing assessment could identify areas of strength and highlight areas where further refinement or expansion is needed.
\end{itemize}

The success of this endeavour hinges on the active participation and contributions of the broader community.
The responsible features presented here are merely an initial step.

Overall, this work complements and synergises with existing responsible music AI initiatives.
While efforts like the Human Artistry Campaign, AI for Music's Principles, and UK Music's policy recommendations provide crucial advocacy and high-level guidance, and other projects like Music RAI and AI:OK address specific concerns such as bias and regulation of music AI (c.f. Introduction), this work differs in at least two ways:
(1) it translates broad ethical principles into actionable features that can be applied during the design and evaluation of generative music AI.
Grounded in the EU's HLEG-AI Guidelines, it aims to provide a structured methodology for operationalising responsible design in generative music AI.
(2) it takes a holistic approach, aiming to serve as an integrating structure.
By establishing a common technical language and a shared vision, it seeks to unify diverse perspectives and previously isolated aspects of generative systems (e.g., transparency, evaluation, data handling) into a cohesive framework for responsible development.

By combining our granular, feature-based approach with the broader ethical principles and policy recommendations of other initiatives, the community can work collectively towards the creation of a just and balanced ecosystem that fosters both innovation in AI and the protection of human artistic expression.
We, therefore, extend an invitation to researchers, developers, musicians, ethicists, legal experts, and all those with a stake in the future of music to engage with these ideas and contribute to shaping a  framework for responsible generative AI in music.

\section{Conclusions}\label{sec:conclusions}

Driven by the transformative rise of generative AI in the creative sector, this work contextualises the EU's HLEG-AI Ethics Guidelines for Trustworthy AI within the domain of music.
We expanded and systematised the framework's seven macro-requirements, resulting in 45 responsible features that aim to guide the development, deployment, and assessment of generation music AI.
By balancing innovation with ethical considerations, we advocate for a tradeoff where artists and AI development collaborate in a way that safeguards, inspires, and augments human creativity and artistry.

This work is distinct in its holistic and actionable approach.
It unifies previously scattered efforts on responsible music AI, translating broad ethical principles into specific design considerations.
We posit that embracing trustworthiness fuels innovation, rather than constraining it.
We discussed how integrating ethical considerations from the outset addresses concerns about data provenance, bias, copyright, and transparency, while simultaneously uncovering new technical challenges and research opportunities to advance generative music as responsible AI.

Crucially, this work is a call for interdisciplinary collaboration.
The identified features aim to establish a shared vocabulary and common ground for AI developers, musicians, ethicists, legal experts, industry, and the public to engage in a productive dialogue.
We propose a roadmap to operationalise these features, emphasising stakeholder involvement, flexible framework implementation, comprehensive evaluation, and semantic documentation.
We also acknowledged the inherent challenges of such implementation: the subjectivity of musical evaluation, the need to align market forces with ethical development, and the fast evolving nature of generative AI, and suggested potential pathways to address these obstacles and mitigate their impact.

Moving forward, research efforts will focus on consolidating our feature set to systematise a framework for responsible generative music AI, leveraging stakeholder collaborations for its continual refinement and extension.
These synergies will also inform the development of tools and platforms to support the framework's utilisation, and to analyse and position existing generative music AI approaches within the landscape defined by these responsible features.

\section{Acknowledgments}

This project has partially received funding from the European Union’s Horizon 2020 research and innovation programme under grant agreements No 101004746 and No 101070412.
This work is also partially supported by the HUMAINT programme (Human Behaviour and Machine Intelligence), Joint Research Centre, European Commission. 
The authors would also like to thank Emilia G\'{o}mez, Eduardo Coutinho, Sabine Jacques, and Tom Brett for their valuable contributions of material and feedback, which significantly improved this work. 

\bibliographystyle{iccc}
\bibliography{iccc}

\appendix
\section{Responsible features for Generative Music AI}\label{appx:features}
The following table provides an overview of the responsible features we have identified through the contextualisation of the HLEG-AI Guidelines for Trustworthy AI \cite{hleg2019ethics} to Generative Music AI.
The feature set presented in this paper is designated as version 1.0 (\texttt{v1.0}).
Each feature belongs to one of the seven macro-requirements for Trustworthy AI (alias \textit{facet}, or \textit{pillar}), whose contextualisation and motivations are reported in Section~2\ref{sec:framework}.
To complement this feature list and facilitate the ongoing consultation and evaluation of these responsible features, a dedicated website is also provided at this link: \RAIMw.
This website serves as a dynamic resource for further exploration and engagement with the framework for responsible generative music AI.

\begin{table*}[t]
\resizebox{\textwidth}{!}{%
{\renewcommand{\arraystretch}{1.5}
\begin{tabular}{@{}rp{.2cm}p{2cm}p{3.0cm}p{10cm}@{}}
\toprule
\multicolumn{1}{l}{\textbf{\#}} &
  \multicolumn{1}{l}{\textbf{R
  }} &
  \textbf{TAI Facet} &
  \textbf{Feature} &
  \textbf{Description} \\ \midrule
1 &
  1 &
  Agency &
  Personalisation &
  The system can reuse the musical repertoire of the user to personalise the style of the generation. \\
2 &
  1 &
  Oversight &
  Creative feedback (HITL) &
  The system can iteratively refine or improve the generation based on the feedback provided by the user throughout the creative process. \\
3 &
  1 &
  Oversight &
  Controllability \quad (conditioning) &
  To steer the generations depending on the user’s preferences, the system provides a rich variety of modalities including language, melodic, harmonic, rhythmic, and emotional control of the generation. \\
4 &
  1 &
  Oversight &
  Red button &
  If the system is generating and playing/displaying music that is either unpleasant, disturbing or contains lyrics with offensive language, the user can always halt the generation process at any time. \\
5 &
  1 &
  Agency &
  Safety disclaimer &
  The system always advises the user whenever it can generate music that may be deemed dangerous or offensive to some categories of users (e.g. offensive language in lyrics). \\
6 &
  1 &
  Agency &
  Deception avoidance &
  The system does not show any deceptive behaviour related to the copyright and ownership of the generated material. For instance, the system never tries to claim ownership or novelty of the generations when music is possibly plagiarised. \\
7 &
  1 &
  Oversight &
  Artist involvement (HOTL) &
  The system has been designed with the active involvement of creative professionals throughout its development cycle. \\ \cmidrule(l){2-5}
8 &
  2 &
  Resilience &
  Music leakage prevention &
  The system does not allow for the full or partial reconstruction of the music material used as training data unless this is explicitly acknowledged and allowed by the copyright holders of the music data. \\
9 &
  2 &
  Fallback plan &
  Generation fallback &
  If the system needs to generate music continuously over time, and the content is considered offensive at some point, the system can switch to a simpler generative strategy (e.g. a rule-based mode). \\
10 &
  2 &
  Accuracy &
  Model evaluation &
  The evaluation of the model/system is consistent with other frameworks and benchmarks for music generation. \\
11 &
  2 &
  Accuracy &
  Music evaluation &
  The system/model provides a comprehensive evaluation of the musical properties of its generations. \\
12 &
  2 &
  Accuracy &
  Expert evaluation &
  The evaluation of the system/model involves creative professionals or music experts. \\
13 &
  2 &
  Reproducibility &
  Model availability &
  The computational model behind the system is fully publicly available and includes pre-trained checkpoints. \\
14 &
  2 &
  Reproducibility &
  Training data availability &
  The training material on which the system relies is fully publicly accessible. \\
15 &
  2 &
  Reproducibility &
  Prompt-to-Gen &
  If sample generations are released, the model can be seeded and prompted to recreate the same musical content. \\
16 &
  3 &
  Privacy &
  Prompt leakage prevention &
  The system neither distributes nor leaks any personal data used by users to prompt the generation of music. \\ \cmidrule(l){2-5}
17 &
  3 &
  Data quality &
  Training metadata integrity &
  The model advises if it is not trained on music data which is fully and correctly attributed to the right authors. \\
18 &
  3 &
  Data quality &
  Safety of training data &
  The model advises if it is trained on music data that can be deemed offensive or socially harmful (e.g. lyrics). \\
19 &
  3 &
  Access to data &
  Prompt governance &
  If the system stores data collected from users through the generation process (e.g., prompts, feedback, music), access to it is fully regulated. \\
\end{tabular}%
}}
\label{tab:features-all-a}
\end{table*}

\begin{table*}[t]
\resizebox{\textwidth}{!}{%
{\renewcommand{\arraystretch}{1.5}
\begin{tabular}{@{}rp{.2cm}p{2cm}p{3cm}p{10cm}@{}}
\toprule
\multicolumn{1}{l}{\textbf{\#}} &
  \multicolumn{1}{l}{\textbf{R}} &
  \textbf{TAI Facet} &
  \textbf{Feature} &
  \textbf{Description} \\ \midrule
20 &
  3 &
  Copyright \& Licensing* &
  Generative reuse of music data &
  The model uses music material (e.g. scores, audio recordings, lyrics, MIDI recordings and transcriptions) whose licensing and terms of use explicitly allow for training systems. \\
21 &
  3 &
  Copyright \& Licensing* &
  Copyright and licensing of generations &
  The system provides guidance or clear and comprehensive information about the copyright and licensing that apply to the generations. \\ \cmidrule(l){2-5}
22 &
  4 &
  Traceability &
  System documentation &
  The system's design is well-documented throughout its development cycle, including instructions for model implementation, training, and data generation. \\
23 &
  4 &
  Traceability &
  Evaluation documentation &
  The evaluation of the model/system is well-documented and reproducible, promoting consistency and transparency in future evaluations (e.g., of other music models). \\
24 &
  4 &
  Traceability &
  Artefact watermarking &
  The system automatically embeds a watermark into every generation to remark on their artificial nature. \\
25 &
  4 &
  Explainability &
  Generation explainability &
  The system can explain how the generations are created in a way that is understandable to its target users. \\
26 &
  4 &
  Explainability &
  Data explainability &
  The system can relate each generation to the training material that contributed to its creation process (e.g. a pattern, motive, or sample). \\
27 &
  4 &
  Communication &
  Artificial awareness &
  Users interacting with the system during the generation process are always aware of its artificial nature. \\
28 &
  4 &
  Communication &
  Benefits communication &
  The benefits of using the particular system, compared to other solutions, are communicated to users prior to its use. \\
29 &
  4 &
  Communication &
  Limitations communication &
  The technical limitations and the potential risks of the system are communicated to users prior to its use. \\
30 &
  4 &
  Communication &
  Instructional material &
  Appropriate instructional material and disclaimers are provided to users on how to adequately use the system. \\ \cmidrule(l){2-5}
31 &
  5 &
  Avoidance of unfair bias &
  Music corpus statistics &
  The system provides a quantification of the kind of music used for training (genre, style, period, etc.), with statistics on the training corpora. \\
32 &
  5 &
  Accessibility \& universal design &
  Accessible interfaces &
  The system provides generative interfaces and/or prompting modalities to make it more accessible and inclusive to users. \\
33 &
  5 &
  Accessibility \& universal design &
  Accessibility assessment &
  If the system provides accessible features, these have been evaluated and tested with the specific target of users they are intended for. \\
34 &
  5 &
  Accessibility \& universal design &
  Accessibility awareness &
  The system explicitly acknowledges whether its use is limited, or not suitable, to certain categories of users. \\
35 &
  5 &
  Stakeholder participation &
  Continuous assessment &
  The system includes music stakeholders (creative professionals, ethical experts, AI engineers and researchers) as part of a long-term strategy for the continuous assessment of its outputs, impact, and trustworthiness. \\ \cmidrule(l){2-5}
36 &
  6 &
  Sustainable \& env. friendly AI &
  Training footprint &
  The system provides an indication of the resources consumed for training the model, in terms of hardware, time, and energy consumption (cost of training). \\
37 &
  6 &
  Sustainable \& env. friendly AI &
  Generation footprint &
  The system provides any indication of the environmental footprint created after generating a whole song or a part of it (cost of inference). \\
\end{tabular}%
}}
\label{tab:features-all-b}
\end{table*}


\begin{table*}[t]
\resizebox{\textwidth}{!}{%
{\renewcommand{\arraystretch}{1.5}
\begin{tabular}{@{}rp{.2cm}p{2cm}p{3cm}p{10cm}@{}}
\toprule
\multicolumn{1}{l}{\textbf{\#}} &
  \multicolumn{1}{l}{\textbf{R}} &
  \textbf{TAI Facet} &
  \textbf{Feature} &
  \textbf{Description} \\ \midrule
38 &
  6 &
  Social impact &
  Responsible data collection &
  If the model uses any human-made annotation (for training, or evaluation), data collection or crowdsourcing was conducted and documented ethically, fairly, and with adequate compensation for annotators. \\
39 &
  6 &
  Social impact &
  Social purpose &
  The system has been designed and used to bring societal benefits, e.g. supporting teaching activities, wellbeing applications, or improving accessibility of creative technologies. \\
40 &
  6 &
  Society \& democracy &
  IP validation &
  The system has mechanisms in place to detect possible cases of plagiarism or IP infringement resulting from the generations. \\
41 &
  6 &
  Society \& democracy &
  Revenue sharing &
  If the system is a paid service, revenues from the generations are also shared with the artists who contributed training data. \\ \cmidrule(l){2-5}
42 &
  7 &
  Auditability &
  Audit access &
  If proprietary, access to the model behind the system and training material can be granted to internal and/or external auditors (on request) and their evaluation reports can be made available. \\
43 &
  7 &
  Minimisation of negative impacts &
  Impact assessment &
  Prior to the deployment of the system, potential negative impacts have been identified, assessed, minimised and openly communicated. \\
44 &
  7 &
  Trade-off &
  Responsible statement &
  The inability of a system to provide responsible features (like those outlined before), in full or in part, is explicitly motivated and documented. Trade-offs should be carefully selected to prioritise the minimisation of risks to ethical principles. \\
45 &
  7 &
  Redress &
  Generative redress &
  When unjust adverse impact occurs (e.g. generations contain offensive content, or the music is plagiarised), the system explicitly accounts for redress (e.g. compensation, direct correction through feedback). \\ 
  &  &  & \vspace{25em} &
  \\ \bottomrule
\end{tabular}%
}}
\caption{List of responsible features for Generative Music AI (\texttt{v1.0}). Legend: \textbf{R} refers to one of the seven Trustworthy AI macro requirements as per above; \textbf{Facet} denotes a specific subcategory of one macro requirement (e.g. ``\textit{Accuracy}'' in Robustness and Safety); \textbf{Feature}, indicates the name of each responsible feature we extracted and defined from the contextualisation of each category; \textbf{Description}, is the outline of the responsible feature in the form of requirement.
}
\label{tab:features-all-c}
\end{table*}

\end{document}